%% file: main.tex
\documentclass[10pt,twocolumn,letterpaper]{article}

\usepackage[pagenumbers]{wacv} 

\definecolor{wacvblue}{rgb}{0.21,0.49,0.74}
\usepackage[pagebackref,breaklinks,colorlinks,allcolors=wacvblue]{hyperref}

\def\wacvPaperID{\ } 
\def\confName{\ }
\def\confYear{\ }

\title{GIBLy: Improving 3D Semantic Segmentation through an Architecture-Agnostic  Lightweight Geometric Inductive Bias Layer}

\author{Diogo Lavado\\
NOVA School of Science and Technology, Portugal and\\
Universit\`a degli Studi di Milano, Italy\\
{\tt\small d.lavado@campus.fct.unl.pt}
\and
Alessandra Micheletti\\
Universit\`a degli Studi di Milano, Italy\\
{\tt\small alessandra.micheletti@unimi.it}
\and
Cl\'audia Soares\\
NOVA School of Science and Technology, Portugal\\
{\tt\small claudia.soares@fct.unl.pt}
}

\begin{document}
\maketitle

\input{sec/0_abstract}

\input{sec/1_intro}
\input{sec/2_related_work}

\input{sec/3_gibli}

\input{sec/4_experiments}

\input{sec/5_conclusions}

{
\small
\bibliographystyle{ieeenat_fullname}
\bibliography{main}
}

\end{document}


\maketitle

\input{sec/6_supp}

{
\small
\bibliographystyle{ieeenat_fullname}
\bibliography{main}
}

%% file: sec/0_abstract.tex
\begin{abstract}
In 3D scene understanding, deep learning models rely on large models and extensive training to capture basic geometric structures that are present in the 3D data. However, existing methods lack explicit mechanisms to incorporate geometric information, such as learnable primitive shapes, often necessitating large models and more training data which in turn increases cost and can limit generalization. 
We introduce \textbf{GIBLy}, a lightweight geometric inductive bias layer that integrates learnable geometric priors into 3D segmentation pipelines. GIBLy enhances existing architectures -- whether MLP-based, convolution-based, or transformer-based -- by providing 
features aligned with simple geometric shapes (and thus human-interpretable) that improve segmentation performance with minimal computational overhead. We validate our approach across multiple 3D semantic segmentation benchmarks, demonstrating consistent performance gains, including up to \textbf{+11.5\% mIoU on TS40K} with PTV3, while adding only \textbf{58K extra parameters}. Our results highlight the benefit of explicitly encoding geometric structure to support accurate and efficient 3D scene understanding, with a lightweight add-on layer
\end{abstract}

%% file: sec/1_intro.tex
\section{Introduction}
\label{sec:intro}
\begin{figure*}[t]
    \centering
    \begin{minipage}{0.55\textwidth}
        \centering
        \includegraphics[width=\linewidth]{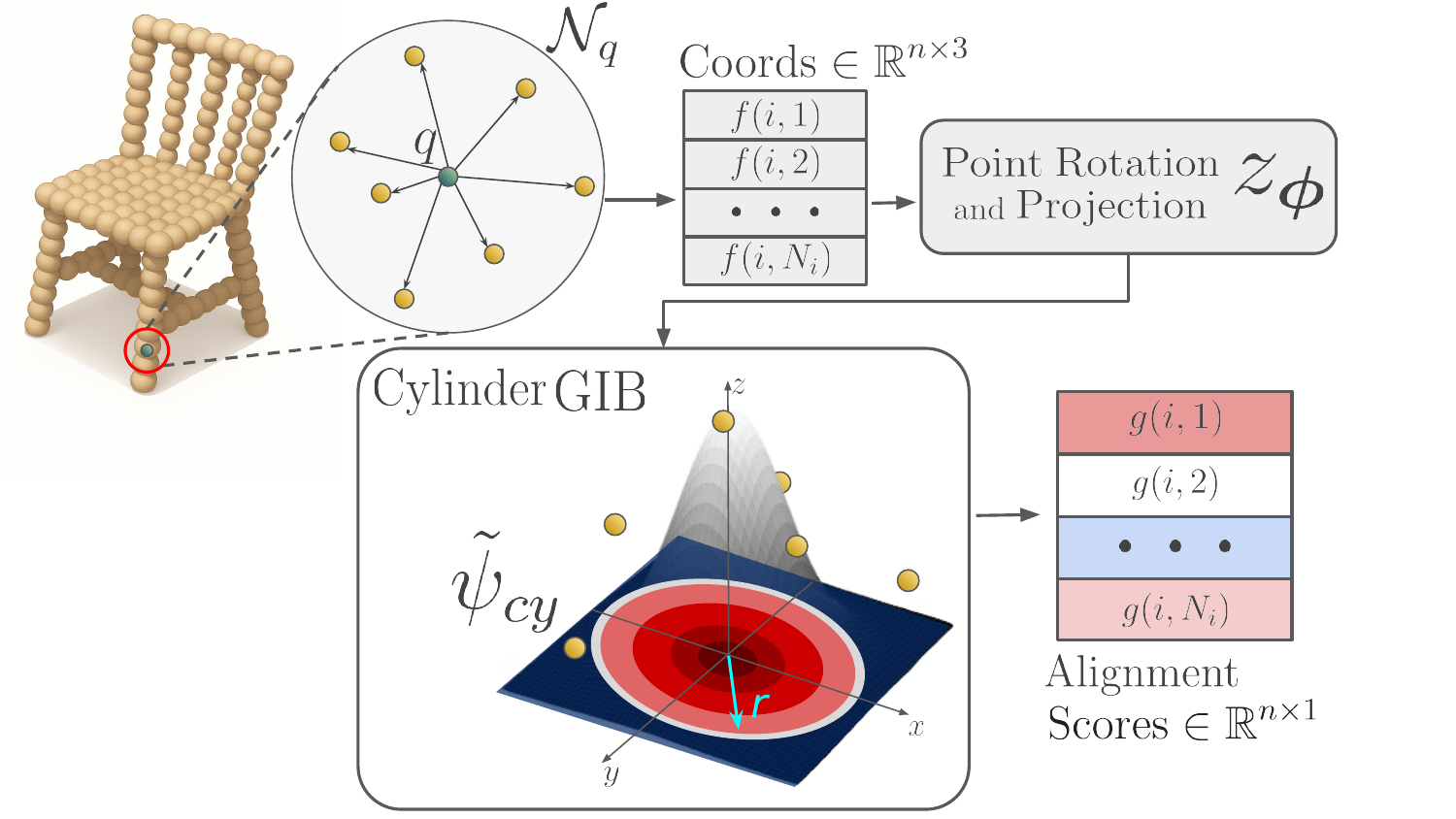}
    \end{minipage}%
    \hfill
    \begin{minipage}{0.44\textwidth}
        \centering
        \includegraphics[width=\linewidth]{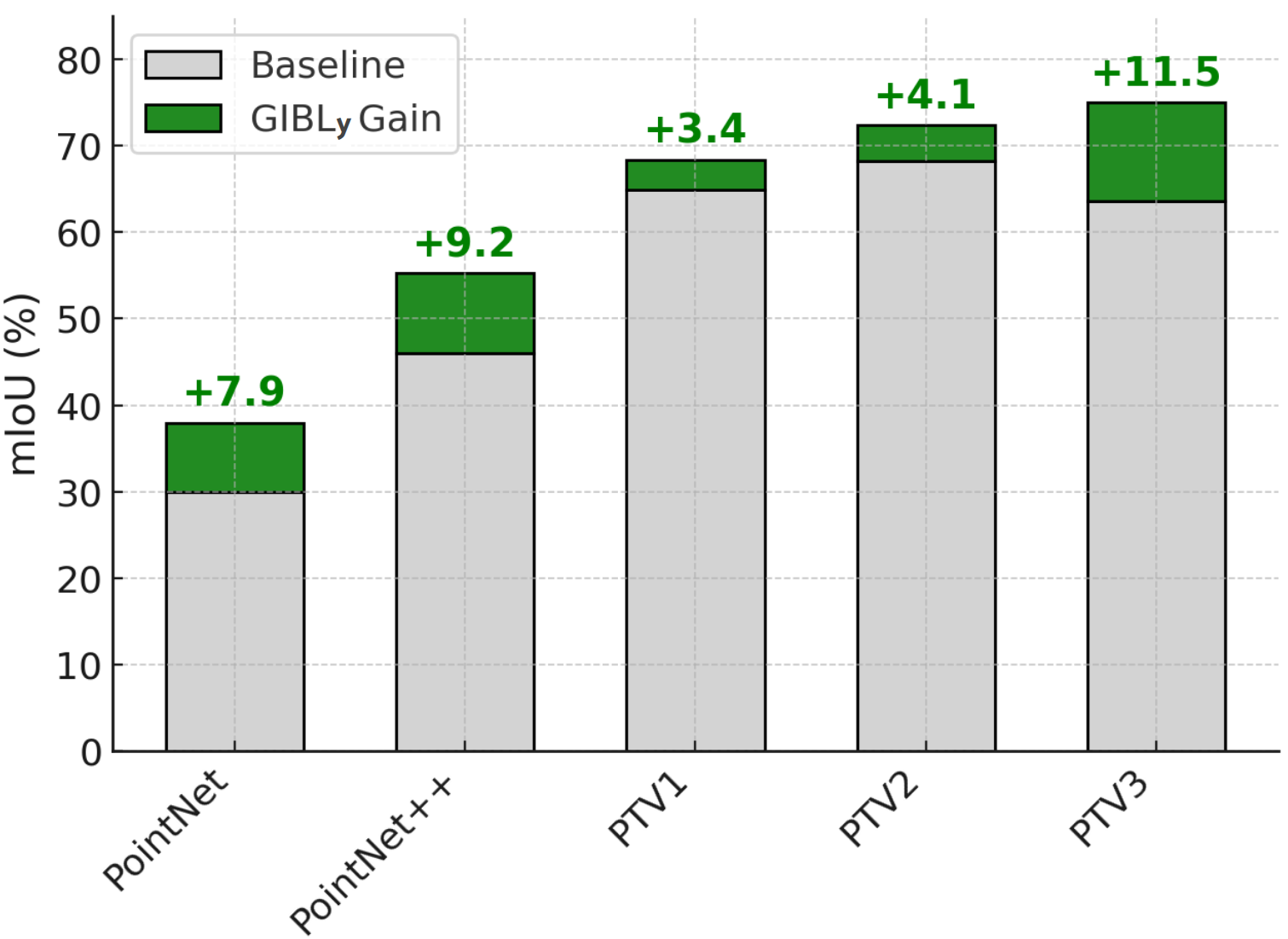}
    \end{minipage}
    \caption{
    \textbf{GIBLy injects learnable geometric priors to improve 3D understanding.}
    \textit{Left}: A Cylinder geometric inductive bias (GIB) aligns to a chair leg (neighborhood $\mathcal{N}_{\text{q}}$), using learned orientation $\boldsymbol{\phi}$ and radius $r$, producing an alignment score.
    \textit{Right}: On TS40K~\cite{lavado2025learning}, adding a GIB-Layer (GIBLy) single-handedly boosts mIoU across multiple backbones (up to \textbf{+11.5\%} mIoU on PTV3~\cite{wu2023ptv3}) with only \textbf{58K extra parameters.}
    }
    \label{fig:teaser}
\end{figure*}

3D data is ubiquitous in applications such as autonomous driving, augmented reality, and robotics. Unlike images, which are organized on regular pixel grids, 3D point clouds lack the fixed pixel array structure: points are scattered in 3D space with varying density, so conventional image CNNs cannot be directly applied.

Current state-of-the-art methods in 3D scene understanding can be broadly categorized by their feature extraction strategies. Early works like PointNet~\cite{qi2017pointnet} and PointNet++~\cite{qi2017pointnet++} employ shared multilayer perceptrons (MLPs) to extract features directly from raw point clouds. Convolution-based approaches, in contrast, leverage the convolution operator to capture both local and global context, either by rasterizing point clouds into volumetric grids~\cite{peng2024oa,maturana2015voxnet,meng2019vv} or by operating directly on the irregular point sets~\cite{thomas2019kpconv,wu2019pointconv,li2018pointcnn}. More recently, transformer-based architectures have emerged~\cite{wang2023octformer,wu2022point,wu2023ptv3,zhao2021point,lai2022stratified}, exploiting attention mechanisms~\cite{vaswani2017attention} to achieve superior point-wise feature extraction and overall performance.
Nonetheless, these methods typically focus on learning geometric relationships implicitly from data, rather than incorporating explicit geometric priors that could guide feature extraction more directly.
They rely on substantial computational resources, memory, and complex architectures to learn geometric relationships that are common across various tasks;
e.g., a network might have to learn from scratch what a \textit{flat surface} or \textit{elongated cylinder} look like, which are concepts that could be built in as priors.
For example, convolution-based methods benefit from built-in inductive biases such as locality and translation equivariance, which have enabled their success in image processing. In contrast, MLP-based methods, including vanilla transformer architectures, lack such inherent biases and must learn spatial relationships entirely from data. 
In 3D, we lack analogous built-in geometric biases for many architectures.
This gap motivates techniques to integrate explicit geometric cues into existing 3D models. By augmenting a model with lightweight, interpretable geometric inductive biases (GIBs), one can enhance feature extraction without modifying the underlying architecture or adding significant computational cost.

To address these limitations, we introduce \textbf{GIBLy}, a lightweight geometric inductive bias layer that integrates learnable parametric geometric priors into 3D deep learning models. GIBLy is model-agnostic and can be integrated without altering the underlying architecture. This layer provides shape-aligned, interpretable features that enhance spatial reasoning without altering the underlying architecture or introducing significant computational overhead.
%
%
We validate GIBLy through extensive experiments across diverse 3D semantic segmentation backbones~\cite{qi2017pointnet,qi2017pointnet++,thomas2019kpconv,zhao2021point,wu2022point,wu2023ptv3} and benchmarks~\cite{lavado2025learning,caesar2020nuscenes,dai2017scannet,armeni20163d,behley2019semantickitti}. Our results show that GIBLy improves performance, including a notable boost of +11.5\% mIoU on TS40K~\cite{lavado2025learning} with PointTransformerV3~\cite{wu2023ptv3}, all while adding just \textbf{58K trainable parameters}.
Our main contributions are:
\vspace{1mm}
\begin{itemize}
    \item We introduce GIBLy, a geometric inductive bias layer that is lightweight, interpretable, and architecture-agnostic. It provides explicit geometric cues for 3D scene understanding and easily integrates into diverse 3D architecture.
    \item We demonstrate that GIBLy improves performance across a wide range of 3D segmentation models and benchmarks with minimal additional parameters and no modifications to the base architecture. Our model can improve mIoU by up to 10\% with only 58K more parameters.
    \item We provide extensive ablation studies showing how the placement, neighborhood size, number of GIB instances, and bias types influence performance and efficiency.
\end{itemize}

%% file: sec/2_related_work.tex
\section{Related Work}
\label{sec:related_work}


\paragraph{3D Scene Understanding.}
In 3D point cloud understanding, deep learning architectures are typically categorized into three paradigms based on their strategy to model point cloud data.
Projection-based methods project 3D point clouds onto different 2D planes to utilize 2D CNN frameworks to infer 3D semantics~\cite{su2015multi,lawin2017deep,yang2019learning,lang2019pointpillars,lyu2020learning}.
Voxel-based methods discretize point clouds as voxel-grids to facilitate 3D convolutions~\cite{maturana2015voxnet,le2018pointgrid,meng2019vv,zhang2020polarnet}. While effective, these voxel approaches face scalability issues; sparse convolution techniques subsequently alleviated this problem~\cite{choy20194d,wang2017cnn,peng2024oa}.
In turn, point-based methods retain the full detail of 3D point clouds by processing them directly~\cite{qi2017pointnet,qi2017pointnet++,li2018pointcnn,thomas2019kpconv,kong2023rethinking}. Recent developments of the attention mechanism~\cite{vaswani2017attention} tailored for point cloud processing have enabled new breakthroughs in performance and are gradually becoming dominant in most prominent benchmarks~\cite{zhao2021point,wu2022point,lai2022stratified,park2022fast,wang2023octformer,lai2023spherical,wu2023ptv3}.
Nevertheless, attention mechanisms suffer from expensive calculation and memory footprint.

\vspace{-6pt}
\paragraph{Convolution-based learning}
%
The lack of structure and irregular density of point clouds sharply contrast with the consistent frame of images.
To circumvent this, projection- and voxel-based methods force point clouds into a structured and rasterized grid, allowing 2D and 3D CNN methods to be employed.
However, 2D projections discard some geometric information of point clouds, and processing 3D voxel-grids incurs high computational and memory costs.

Graph convolutional networks in the spatial domain offer a natural alternative for feature extraction by interpreting point clouds as geometric graphs~\cite{simonovsky2017dynamic,hermosilla2018monte}. 
These strategies are typically divided by the nature of their convolution kernels:
discrete-kernel methods~\cite{lei2019octree,thomas2019kpconv,lei2020spherical,zhu2021cylindrical,wang2019graph} follow standard CNN kernels and define filters on regular grids in the euclidean space, similar to 3D-CNN kernels for voxel-grids~\cite{maturana2015voxnet}.
In turn, continuous-kernel methods~\cite{groh2018flex,wang2018deep,xu2018spidercnn,li2018pointcnn,wu2019pointconv} parameterize convolution as a function of local point coordinates. For instance, these techniques may represent the filter as $w = h(x_j - x_i)$, where $h$ is a continuous function and $x_j$ is a neighbor of $x_i$.
Typically, $h$ is expressed as a multi-layer perceptron (MLP) due to their theoretical ability to approximate any continuous function.
In this work, we augment the convolution kernel function $h$ with a continuous parametric function that represents a geometric inductive bias. 
This geometric bias serves as an interpretable feature extractor, providing prior knowledge to the network and aiding convergence.

\vspace{-6pt}
\paragraph{Geometric Informed Learning.}
Incorporating geometric information into neural networks has been extensively studied in learning-based surrogate models for solving partial differential equations (PDEs) in fluid dynamics~\cite{li2024geometry,oldenburg2022geometry,cameron2024nonparametric}, as well as in object generation using meshes~\cite{hanocka2019meshcnn} or enforcing shape constraints~\cite{berzins2024geometry}.
In point cloud processing, prior works~\cite{bocchi2022geneonet,lavado2023low,lavado2024scene} have leveraged geometric priors for various tasks. 
These methods demonstrate the benefit of geometric priors, but each is specific to a domain and not general-purpose, such as detecting protein pockets~\cite{bocchi2022geneonet} and identifying power grid components~\cite{lavado2024scene}.
In this work, we capitalize on the fact that 3D objects can often be described by basic geometric properties to develop a general-purpose geometric feature extractor model. Additionally, we demonstrate how integrating explicit geometric features into existing architectures can enhance performance across diverse tasks.

\vspace{-6pt}
\paragraph{Inductive biases in 3D scene understanding.}
%
Inductive biases are essential for effective learning in computer vision. As formalized by the no-free-lunch theorem~\cite{baxter2000model,goyal2022inductive}, no model can generalize well across tasks without incorporating task-specific assumptions.
In 3D scene understanding, geometric inductive biases are especially valuable due to the structured nature of real-world environments. Encoding these biases aligns the learning process with common geometric patterns, improving scene interpretation and reconstruction~\cite{yifan2022input}.
Convolutional networks benefit from built-in biases like locality and translation equivariance~\cite{cohen2016inductive,wang2023theoretical}, which support spatial structure learning. In contrast, MLPs lack such priors and must learn spatial relationships entirely from data. Augmenting MLPs with equivariant mechanisms has been shown to improve 3D understanding~\cite{sun2021canonical}. Similarly, transformer-based models exhibit inherent symmetries~\cite{lavie2024towards}, and benefit from explicit geometric priors in applications such as depth estimation~\cite{xu2023se} and split training~\cite{xu2024permutation}.
In this work, we incorporate geometric priors directly into existing 3D networks via GIBLy, a lightweight layer that improves performance without changes to the original network architecture.

%% file: sec/3_gibli.tex
\section{Geometric Inductive Bias Layer}
\label{sec:GIBLy}

\begin{figure*}[t]
    \centering
    \begin{subfigure}[c]{0.55\textwidth}
        \centering
        \includegraphics[width=\linewidth]{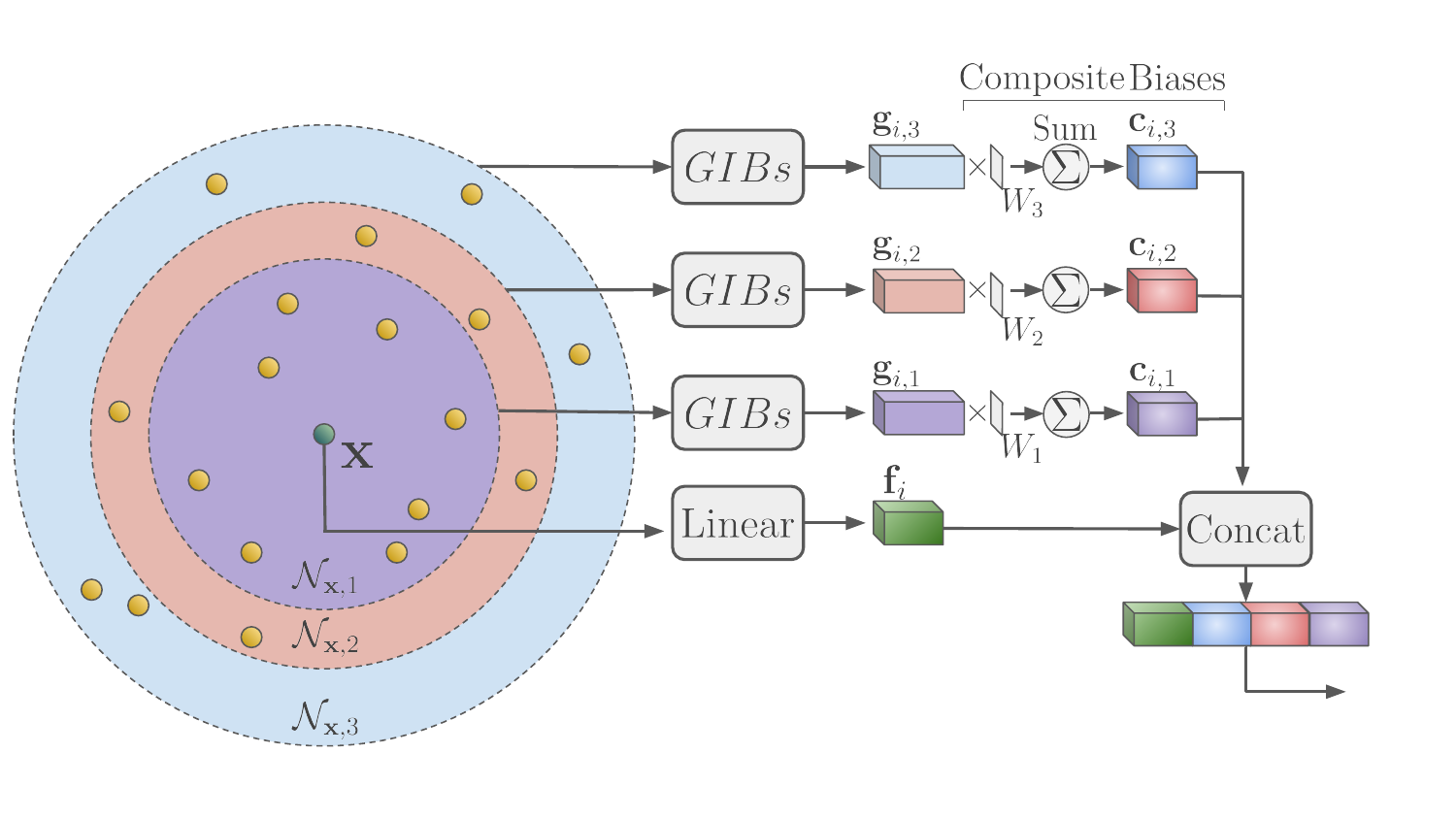}
        \caption{GIBLy module}
        \label{fig:gibly-module}
    \end{subfigure}
    \hfill
    \begin{subfigure}[c]{0.43\textwidth}
        \centering
        \includegraphics[width=\linewidth]{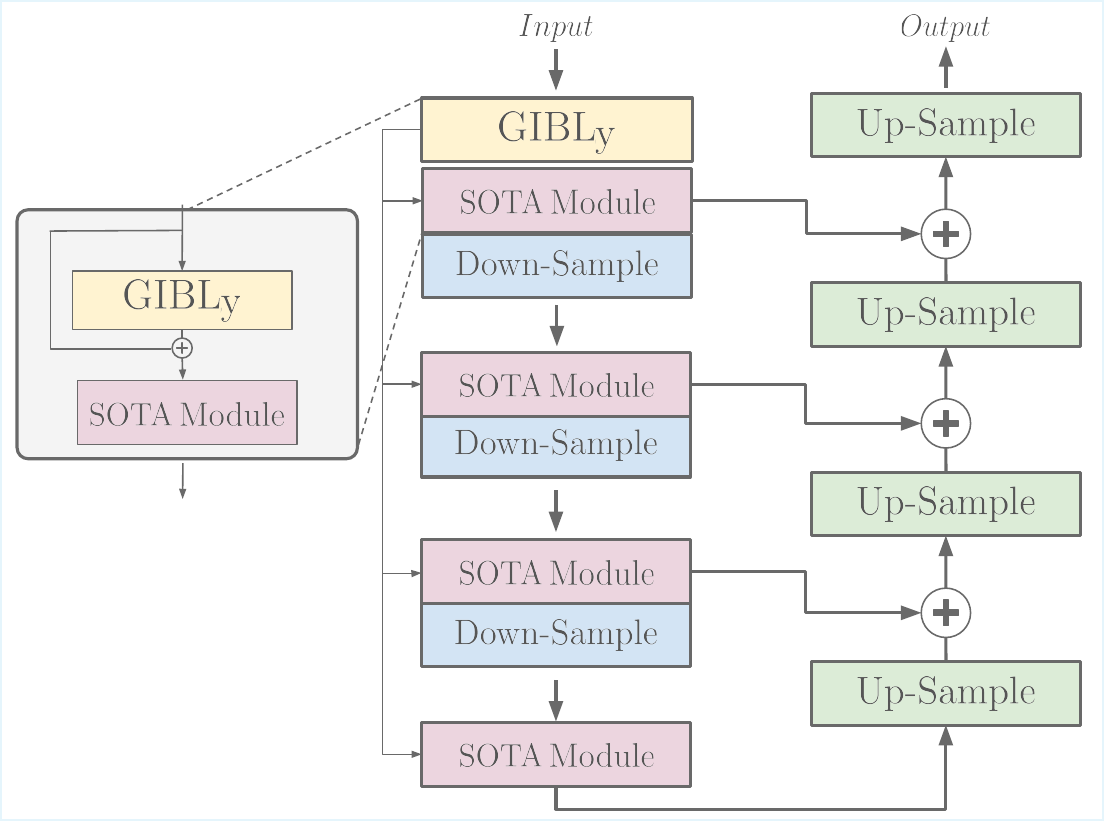}
        \caption{Integration within SOTA architectures}
        \label{fig:gibly-net}
    \end{subfigure}
    \vspace{-5pt}
    \caption{\textbf{Overview of our method.} 
    \textit{(a)} Our GIBLy module injects geometric awareness into point features by applying a set of learnable geometric inductive biases (GIBs) at each query point. We adopt a multi-region design: all input points are treated as query points, and multiple neighborhood scales are considered per point. For a given query $i$, $\textbf{g}_{i,N}$ denotes the GIB alignment scores computed at $N$ neighborhoods. These are linearly combined via learned weights $W_N$ to produce composite bias responses $\textbf{c}_{i,N}$, capturing richer geometric structure. The weights in $W_N$ reflect the importance of each bias. The resulting features are concatenated with the original input features through MLPs or projection layers to produce enriched point representations.
    \textit{(b)} GIBLy is applied once at the input level of a standard 3D backbone, where the geometry of the point cloud is most faithful. Its output features are propagated via skip connections to every downsampling stage in the encoder, allowing geometric information to be reused efficiently without reapplying the GIBLy module at every level.
    }
    \label{fig:gibly-overview}
\end{figure*}

In this section we present the Geometric Inductive Bias Layer (GIBLy).
The goal of GIBLy is to enrich 3D point features with explicit geometric structure while remaining lightweight and modular.
First, we describe the general architecture of the layer in~\cref{subsec:arch}. Next, we introduce a family of geometric inductive biases (GIBs) and their general form in~\cref{subsec:gibs}. Lastly, we present the composition of GIBs into composite biases in~\cref{subsec:observers}.

\paragraph{Motivation.}
\label{subsec:motivation}

A central challenge in 3D scene understanding is that networks must learn geometric concepts directly from raw point clouds, which are irregular, sparse, and often incomplete. Unlike images, where convolutions benefit from built-in inductive biases, 3D backbones lack explicit geometric priors and must infer basic patterns (e.g., cylinders, planes) from data, often requiring large models and training.  
Yet, most objects can be decomposed into simple primitives: a table into a plane and four cylinders, or a car into an ellipsoid and narrow cylinders. While such structures recur across categories, their variation in design, scale, and shape makes fixed handcrafted priors too rigid~\cite{bocchi2022geneonet,lavado2023low}.  
To address this, we propose GIBLy, a lightweight and modular layer that equips 3D backbones with learnable shape-aligned priors, injecting explicit geometric cues as point features at the input stage.

\subsection{GIB-Layer (GIBLy)}
\label{subsec:arch}

GIBLy is designed to inject explicit, interpretable geometric structure into point features while being simple to integrate into existing backbones (Fig.~\ref{fig:gibly-overview}).  
Each input point $q$ is treated as a query, and multiple neighborhoods $\mathcal{N}_q$ at different radii capture geometry across scales. Within each neighborhood, learnable geometric inductive biases (GIBs) measure the alignment of neighbors with primitives such as cylinders, cones, and disks, producing alignment scores $g(x)$.  
To move beyond simple shapes, GIBs are linearly combined into composite biases $\gamma$, enabling richer yet interpretable patterns. Learned coefficients weight the contribution of each primitive, highlighting which biases are most influential.  
The resulting geometric cues are concatenated with original point features, yielding enriched representations that can be processed by downstream modules. Crucially, GIBLy is applied only once at the input stage, where geometry is most reliable, enriched features are then propagated via skip connections through the network, avoiding repeated and less stable recomputation after downsampling.

\subsection{Geometric Inductive Biases (GIBs)}
\label{subsec:gibs}

\paragraph{GIBs on raw point clouds.}
To assess how well a neighborhood of points conforms to a geometric primitive, we require an operator capable of aggregating unordered local point sets and processing them with a tailored kernel. Point-based convolution naturally provides this functionality: it evaluates neighbors relative to a query point while allowing for the use of specialized kernels. 
Accordingly, geometric inductive biases (GIBs) can be integrated into raw point clouds by modeling them as geometric graphs and applying point-based convolutions. This operator is particularly well suited for GIBs because it is translation-equivariant, ensuring that geometric features are detected independently of their absolute position in space.

\vspace{-8pt}
\paragraph{GIBs as point-based convolutions.}
A raw point cloud is generally represented as $\mathcal{P} \in \mathbb{R}^{N\times(3 + C)}$, where $N$ is the number of points, and $3 + C$ denotes the spatial coordinates along with any additional point-wise features, such as color or intensity. Point-based continuous convolution treats $\mathcal{P}$ as a sampling of an underlying continuous function $f: \mathbb{R}^3 \mapsto \mathbb{R}^C$,
with a kernel function $g : \mathbb{R}^3 \mapsto \mathbb{R}$ given by:
\begin{align}
    (\mathcal{P} \ast g)(\textbf{x}) &= \int_{\mathbb{R}^3} g(\textbf{y} - \textbf{x})f(\textbf{y}) d\textbf{y} \\
                              &\approx 
                              \sum_{\textbf{y} \in \mathcal{N}_{\textbf{x}}} g(\textbf{y} - \textbf{x}) f(\textbf{y}),
\end{align}  
where $\mathcal{N}_{\textbf{x}}$ denotes the local neighborhood of $\textbf{x}$.  
Continuous-convolution approaches~\cite{xu2018spidercnn,li2018pointcnn,wu2019pointconv} typically approximate the kernel $g$ using a multi-layer perceptron (MLP) due to its universal approximation theorem. However, such MLP-based methods must learn spatial relationships from data, which can increase model complexity, slow convergence, and reduce robustness to novel inputs.
To alleviate these issues, we incorporate a family of continuous functions that provide alignment scores informed by geometric priors. Specifically, we parameterize $g$ as:  
\begin{equation}
    g(\textbf{z}) = \psi(\textbf{z}; \eta),
\end{equation} 
where $\textbf{z} = \textbf{y} - \textbf{x}$ represents the projection of a neighbor point $\textbf{y}$ onto a canonical reference space, $\psi$ is a geometric basis function (e.g. radial basis function) that encodes specific shape attributes, with $\eta$ serving as a learnable parameter that defines these shape characteristics (e.g., the radius of a cylinder). 
This formulation enables the network to efficiently capture and utilize geometric structures while preserving a transparent interpretation of the shape parameters $\eta$.
%

\vspace{-6pt}
\paragraph{Designing GIBs.}
%
%
Each GIB is implemented as a radial basis function (RBF) defined in terms of the relative position of neighboring points. 
Specifically, given a query point $q$ and its local neighborhood $\mathcal{N}_q$, we define the position of each neighbor $x \in \mathcal{N}_q$ relative to $q$ (i.e., we center each neighbor at the query point). This relative vector is then rotated by a learnable rotation matrix $R_{\boldsymbol{\phi}}$ and projected into a reference space:
\begin{equation}
    z_{\boldsymbol{\phi}}(x) = z(R_{\boldsymbol{\phi}}^\top (x - q)), \quad g(x) = \psi_g(z_{\boldsymbol{\phi}}(x); \boldsymbol{\eta}),
\end{equation}
where $\psi_g$ is a geometric kernel (e.g., Gaussian), $\boldsymbol{\phi} = [\phi_x, \phi_y, \phi_z]$ are learned rotation angles, and $\boldsymbol{\eta}$ are shape parameters (e.g., radius, thickness). 
The vector $\boldsymbol{\phi}$ explicitly parametrizes the rotation matrix $R_{\boldsymbol{\phi}}$, enabling the model to learn a rotation that aligns the relative neighbor coordinates with the GIB $\psi_g$, thus providing a learnable rotation equivariance per GIB instead of enforcing a predefined rotation equivariance.
Consequently, this learned rotation effectively orients the geometric cues encoded by $\psi_g$, allowing the GIB to adapt to different local geometric configurations.
The output $g(x)$ denotes the alignment score of the neighbor $x$ with respect to the geometric prior centered on $q$.
Real-world objects seldom adhere to strict geometric forms, making fixed boundaries unsuitable for capturing their variability. Thus, RBFs offer a continuous similarity measure between points in a local neighborhood and the encoded geometric priors. This enables a more adaptive representation of natural structures without enforcing hard constraints.
We selected the primitives cylinders, cones, planes/disks, ellipsoids as basic components to common structures in man-made and natural environments, such as columns, slopes, and blobs.
We now introduce the Cylinder GIB and its hollowed variant, which serve as representative examples of our geometric bias formulation. To avoid repetition, detailed definitions of the remaining GIB types, including cones, disks, and ellipsoids, are provided in the Supplementary Material.

\vspace{-8pt}
\paragraph{Cylinder GIB.}
The Cylinder GIB models long, tubular structures that are common in both indoor (e.g., table legs, poles) and outdoor (e.g., tree trunks, signposts) scenes. It uses a Gaussian function centered on a cylindrical axis, favoring radial symmetry around a central spine:
\begin{equation}
\psi_{cy}(x) = \exp\left(-\frac{\|z_{\boldsymbol{\phi}}(x)\|^2}{2r^2}\right)
\end{equation}
Here, $z_{\boldsymbol{\phi}}(x)$ represents the projection of point $x$ into a canonical coordinate system defined by the learned rotation matrix $R_{\boldsymbol{\phi}}$, which aligns the cylinder with relevant features in the scene. The term $\|z_{\boldsymbol{\phi}}(x)\|$ corresponds to the radial distance from the cylinder’s central axis.
The parameter $r$ defines the cylinder radius and controls the spread of the Gaussian kernel. Points close to the axis receive high alignment scores, while those farther away are suppressed. The shape parameters for this GIB are $\eta_{cy} = [r, \boldsymbol{\phi}]$ and a depiction of these operations can be found in~\cref{fig:teaser} on the left. 

\vspace{-8pt}
\paragraph{Hollow Cylinder GIB.}
In many real-world scenarios, especially with LiDAR or RGB-D data, cylindrical structures may exhibit hollow interiors due to surface-only sampling. Examples include pipes, cables, or hollow rods. To model these shell-like geometries, we define the Hollow Cylinder GIB:
\begin{equation}
\psi_{hcy}(x) = \exp\left(-\frac{(\|z_{\boldsymbol{\phi}}(x)\| - r)^2}{2t^2}\right)
\end{equation}
Unlike the standard cylinder, this function peaks at a fixed distance $r$ from the axis, suppressing the core and exterior. The parameter $t$ controls the shell thickness, effectively defining the tolerance band around the ring. 
The shape parameters are extended to $\eta_{hcy} = [r, t, \boldsymbol{\phi}]$.

\begin{figure}[t]
    \centering
    \includegraphics[width=\columnwidth]{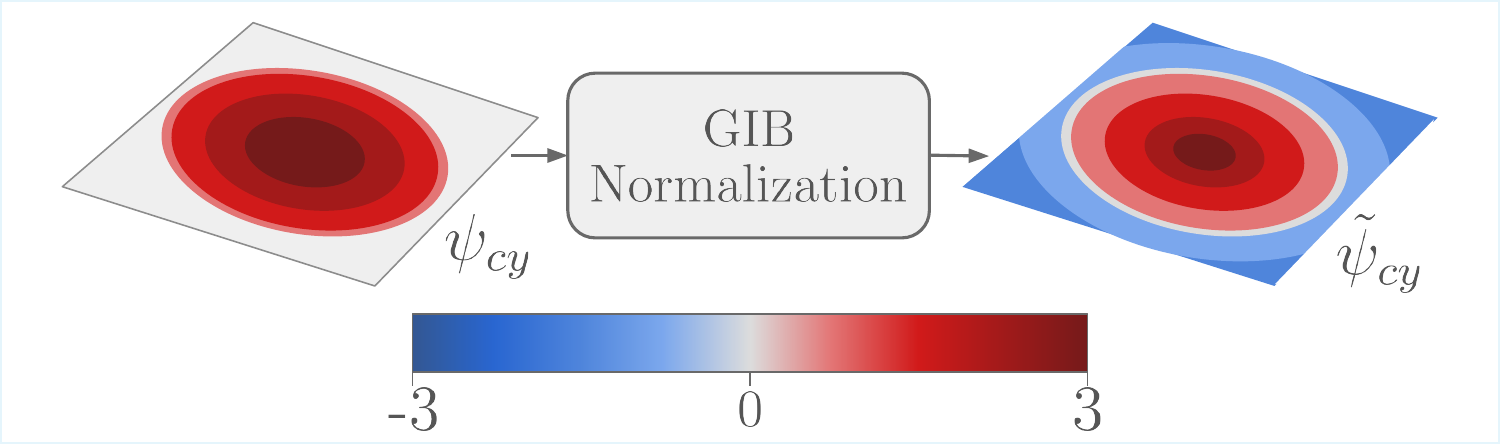}
    \caption{A normalized bias ensures that neighbors aligning with the desired geometric configuration receive positive values, while those deviating receive negative values. Here, a Cylinder GIB is normalized such that neighbor points receive an alignment score congruent with their proximity to the center. Once the radius is exceeded, points fall outside the geometric bias and receive negative scores, indicating deviation from the intended shape.}
    \label{fig:gib_normalization}
\end{figure}

\vspace{-5pt}
\paragraph{GIB normalization.}
Similar to batch normalization layers in convolution-based architectures, we propose a normalization operation for geometric inductive biases. In conventional CNNs, batch normalization is applied directly to the feature maps produced by convolutional layers to standardize their distributions across a mini-batch. However, due to the distinct nature of GIB modules, a direct application of batch normalization is not feasible seeing as it could invert the semantics of alignment scores (i.e., assigning negative values to geometrically aligned neighbors and positive values to misaligned ones).
For a given GIB instance $ \psi$, its score over a local neighborhood $ \mathcal{N}_q $ quantifies the degree to which each neighbor is aligned with the geometric pattern encoded by $ \psi$. Consequently, the normalized bias, denoted by $ \tilde{\psi}$, is designed to yield positive values for neighbors that conform to the desired geometric configuration and negative values for those that deviate. 
This normalization not only ensures that the alignment scores are on a comparable scale across different neighborhoods but also reinforces the intended geometric characteristics while suppressing spurious or misaligned shapes.
To this end, we employ a Monte Carlo approximation to estimate the integral of $ \psi $ over the neighborhood domain $ \mathcal{N}_q $. Specifically, we define the cumulative alignment score as:
\begin{equation}
    \Omega(x) = \int_{\mathcal{N}_q} \psi(y)dy \approx \sum_{y \in \mathcal{MC }} \psi(y).
\end{equation}
Here, the set $ \mathcal{MC} = \{ y \in \mathbb{R}^3 \mid \|y\| \leq r \}$ serves as a Monte Carlo approximation of the neighborhood $\mathcal{N}_q$, where $r$ denotes the neighborhood's radius. The normalized GIB is computed by subtracting the average alignment over the neighborhood from the original score:
\begin{equation}
    \tilde{\psi}(x) = \psi(x) - \frac{\Omega(x)}{|\mathcal{MC}|}.
\end{equation}
Normalizing GIBs is essential for robustly integrating geometric cues into deep learning frameworks. This operation is illustrated in~\cref{fig:gib_normalization}. 

\subsection{Composite Biases}
\label{subsec:observers}
Although geometric inductive biases (GIBs) offer several degrees of freedom via their shape parameters $\eta$ to adapt to input data, a single GIB may be insufficient to capture the intricate details inherent in sophisticated objects. 
To address this limitation, we propose a framework in which multiple GIB instances are combined into composite biases through linear combinations.
Specifically, let $\Psi = \{\psi_j\}_{j=1}^m$ denote a set of $m$ GIBs, where each instance $\psi_j$ is selected from a parametric family belonging to 
$\{\psi_{cy},\, \psi_{cn},\, \psi_{ellip},\, \psi_{dk}, \psi_{hcy},\, \psi_{hcn},\, \psi_{hellip},\, \psi_{hdk}\}.$ 
These biases are aggregated via a linear combination with a learnable weight matrix $W \in\mathbb{R}^{n \times m}$, where $n$ is the number of composite biases, collected in the set $\Gamma = \{\gamma_i\}_{i=1}^n$. Formally, each composite bias is defined as
\begin{equation}
    \gamma_i = \sum_{\psi_j \in \Psi} W_{ij} \, \psi_j.
\end{equation}
In this formulation, each row $W_i$ quantifies the contribution of the individual GIBs to the composite bias $\gamma_i$. A positive coefficient $W_{ij}$ indicates that the corresponding GIB $\psi_j$ contributes constructively, activating the detection of specific patterns within the scene. 
In contrast, a negative coefficient implies that $\psi_j$ exerts a suppressive influence, either altering the resulting shape or diminishing the detection of certain features.
Moreover, the absolute value $|W_{ij}|$ serves as an indicator of the intensity and contribution of the $j$-th inductive bias within the model. A larger magnitude indicates that the model assigns greater importance to that GIB instance in forming the feature representation.
It is important to note that this linear combination approach may introduce unwanted noise into composite biases, due to residual contributions from GIBs that are not pertinent to a particular composite $\gamma_i$. To mitigate this issue, we promote sparsity in the weight matrix by employing an $L_1$ penalty, and further regulate the overall magnitude of the weights through an $L_2$ penalty to prevent overfitting.

%% file: sec/4_experiments.tex
\section{Experiments}
\label{sec:experiments}

\begin{table*}[t]
\centering
\caption{Semantic segmentation results on TS40K. GIBLy improves class-wise and mean IoU across most backbones. Gains are shown in \textcolor{ForestGreen}{green} and deficits in \textcolor{BrickRed}{red}.}
\label{tab:ts40k}
\begin{tabular}{llccccccc}
\toprule
\textbf{Method} & & \textbf{mIoU (\%)} & Noise & Ground & Low Veg. & Mid Veg. & Tower & Power Line \\
\midrule
KPConv & base & 52.77 & 57.02 & 64.75 & 37.12 & 34.63 & 37.36 & 89.99 \\
       & + GIBLy & 46.83 & 53.37 & 61.96 & 35.73 & 31.26 & 12.91 & 85.78 \\
       & $\Delta$ & \textcolor{BrickRed}{--5.94} & \textcolor{BrickRed}{--3.65} & \textcolor{BrickRed}{--2.79} & \textcolor{BrickRed}{--1.39} & \textcolor{BrickRed}{--3.37} & \textcolor{BrickRed}{--24.45} & \textcolor{BrickRed}{--4.21} \\
\midrule
PTV1 & base & 64.90 & 57.50 & 77.33 & 60.34 & 46.51 & 54.19 & 93.54 \\
     & + GIBLy & 68.33 & 63.88 & 79.15 & 68.77 & 50.93 & 52.16 & 95.06 \\
     & $\Delta$ & \textcolor{ForestGreen}{+3.43} & \textcolor{ForestGreen}{+6.38} & \textcolor{ForestGreen}{+1.82} & \textcolor{ForestGreen}{+8.43} & \textcolor{ForestGreen}{+4.42} & \textcolor{BrickRed}{--2.03} & \textcolor{ForestGreen}{+1.52} \\
\midrule
PTV2 & base & 68.29 & 61.16 & 80.13 & 68.17 & 51.39 & 54.48 & 94.43 \\
     & + GIBLy & 72.35 & 67.79 & 82.40 & 73.17 & 54.38 & 60.29 & 96.06 \\
     & $\Delta$ & \textcolor{ForestGreen}{+4.06} & \textcolor{ForestGreen}{+6.63} & \textcolor{ForestGreen}{+2.27} & \textcolor{ForestGreen}{+5.00} & \textcolor{ForestGreen}{+2.99} & \textcolor{ForestGreen}{+5.81} & \textcolor{ForestGreen}{+1.63} \\
\midrule
PTV3 & base & 63.55 & 59.23 & 70.77 & 50.47 & 43.86 & 61.42 & 95.53 \\
     & + GIBLy & 75.03 & 68.91 & 82.96 & 73.32 & 55.33 & 72.49 & 97.17 \\
     & $\Delta$ & \textcolor{ForestGreen}{+11.48} & \textcolor{ForestGreen}{+9.68} & \textcolor{ForestGreen}{+12.19} & \textcolor{ForestGreen}{+22.85} & \textcolor{ForestGreen}{+11.47} & \textcolor{ForestGreen}{+11.07} & \textcolor{ForestGreen}{+1.64} \\
\bottomrule
\end{tabular}
\end{table*}

\subsection{Implementation Details}

\paragraph{Baselines.}
We conducted experiments integrating our proposed GIBLy method into several 3D learning pipelines, including the seminal MLP-based methods PointNet~\cite{qi2017pointnet} and PointNet++~\cite{qi2017pointnet++} (in Supplementary Material), the convolution-based method KPConv~\cite{thomas2019kpconv}, and the transformer-based PointTransformer architectures~\cite{zhao2021point,wu2022point,wu2023ptv3}.  
GIBLy was evaluated on standard benchmarks for indoor and outdoor 3D semantic segmentation. For indoor scene understanding, we consider ScanNet v2~\cite{dai2017scannet} and the S3DIS dataset~\cite{armeni20163d}. 
%
For outdoor scene segmentation, we evaluate on three benchmarks: nuScenes~\cite{caesar2020nuscenes}, SemanticKITTI~\cite{behley2019semantickitti}, and TS40K~\cite{lavado2025learning}. 
%
While we evaluate across all datasets, we adopt TS40K as our primary benchmark for ablations and analysis. Unlike driving datasets, TS40K provides high-resolution LiDAR scans with dense sampling, no occlusion, and anisotropic point distributions within the same object. These properties allow neighborhoods to be reliably estimated and make geometric effects more directly observable, which is essential for isolating the contribution of GIBLy. The dataset also contains objects with strong global structure but diverse local variations (e.g., power line towers), which aligns well with the strengths of radial basis functions in capturing smooth global alignment while tolerating some local variation. Importantly, TS40K’s released configurations allowed us to reproduce baselines faithfully, enabling controlled and reproducible ablation studies.

\subsection{Evaluation}

\begin{figure}[t]
    \centering
    \setlength{\tabcolsep}{1pt}
    \renewcommand{\arraystretch}{1.0}
    \begin{tabular}{ccc}
        \includegraphics[width=0.31\columnwidth]{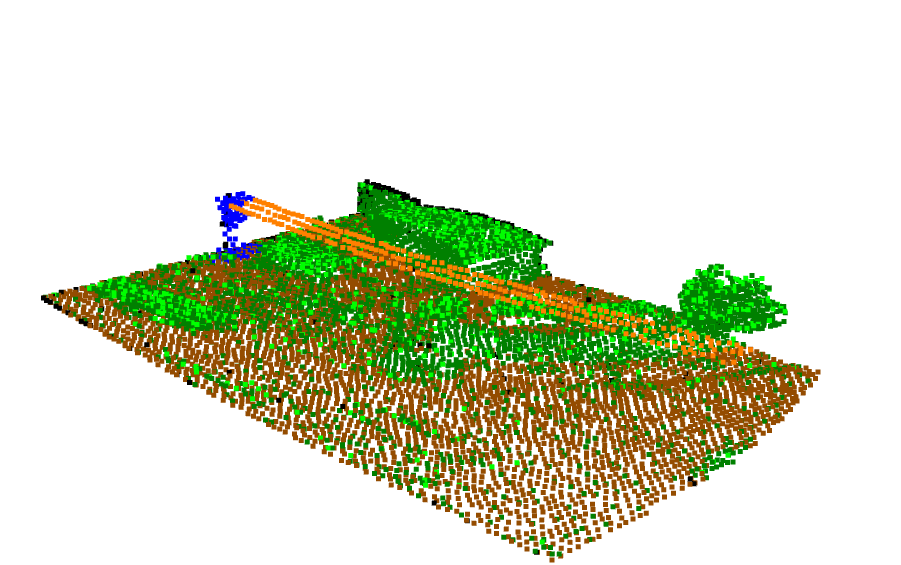} &
        \includegraphics[width=0.31\columnwidth]{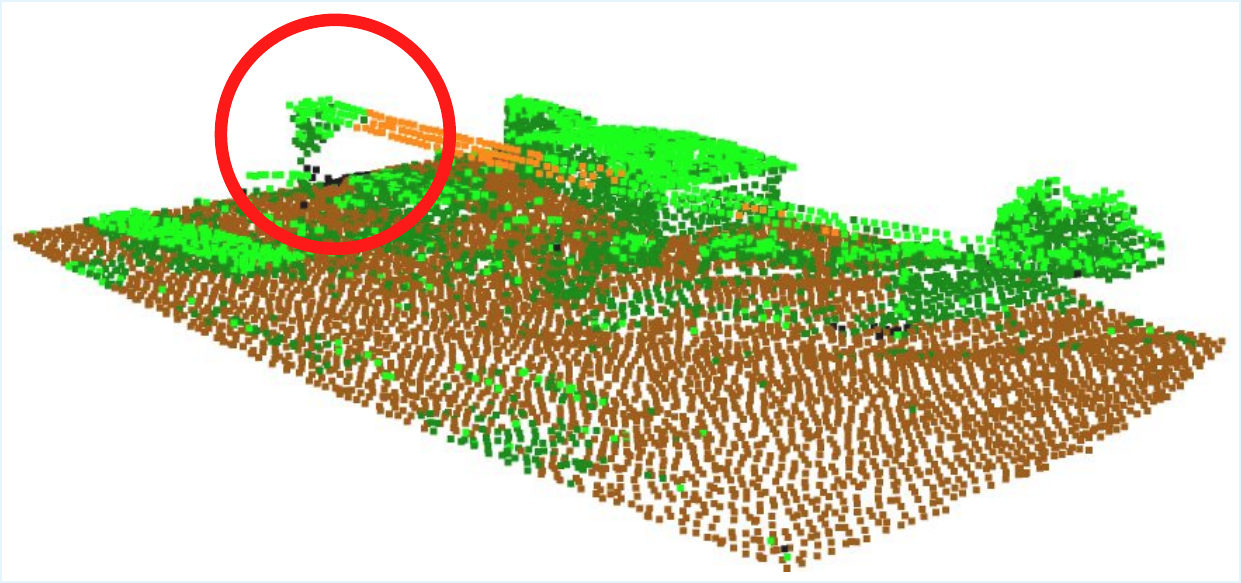} &
        \includegraphics[width=0.31\columnwidth]{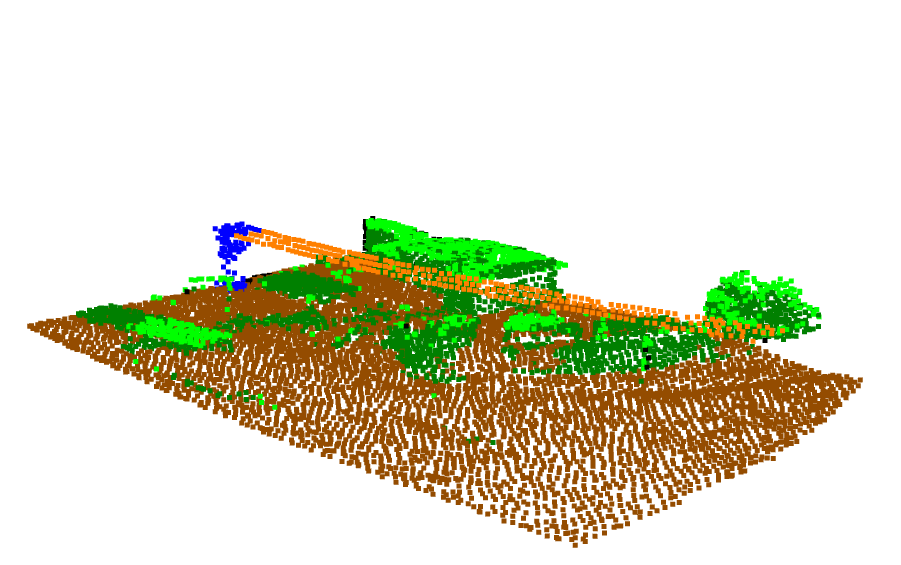} \\
        \includegraphics[width=0.31\columnwidth]{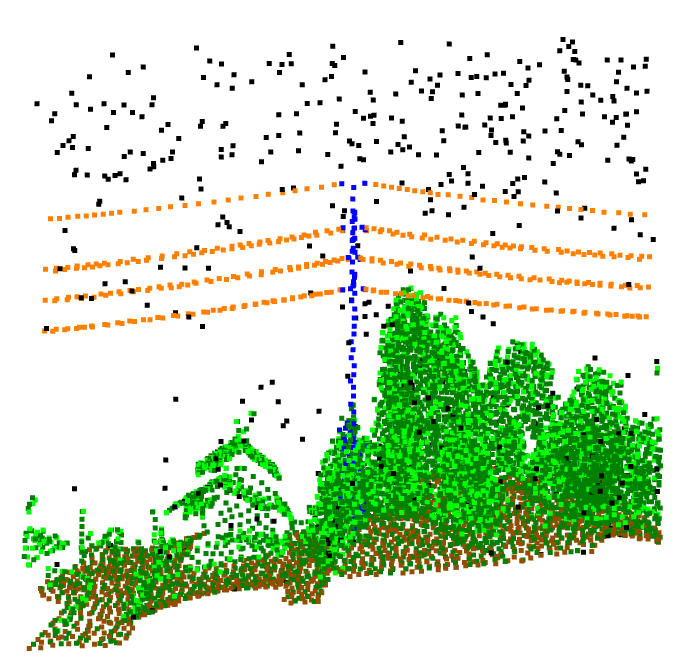} &
        \includegraphics[width=0.31\columnwidth]{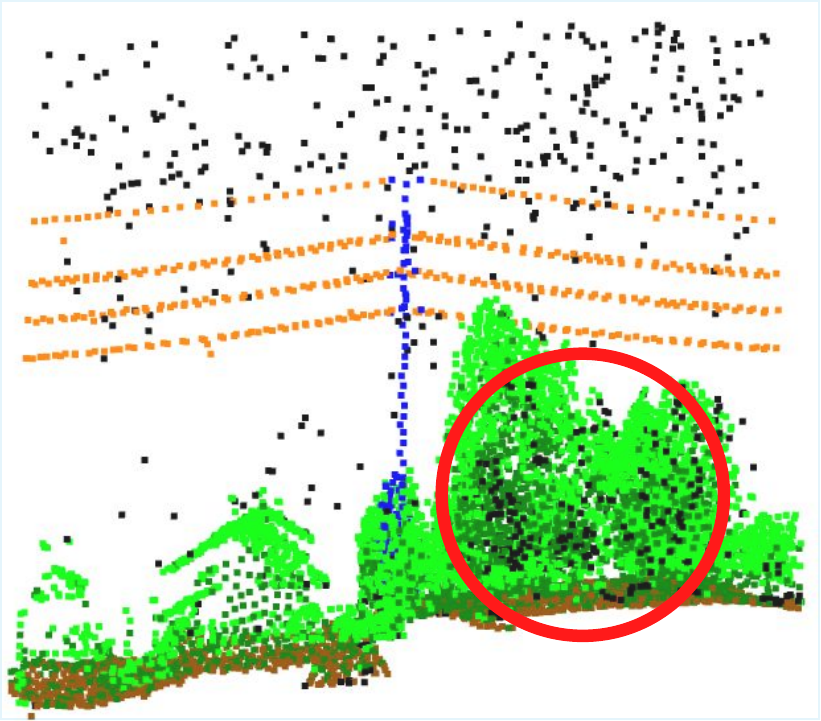} &
        \includegraphics[width=0.31\columnwidth]{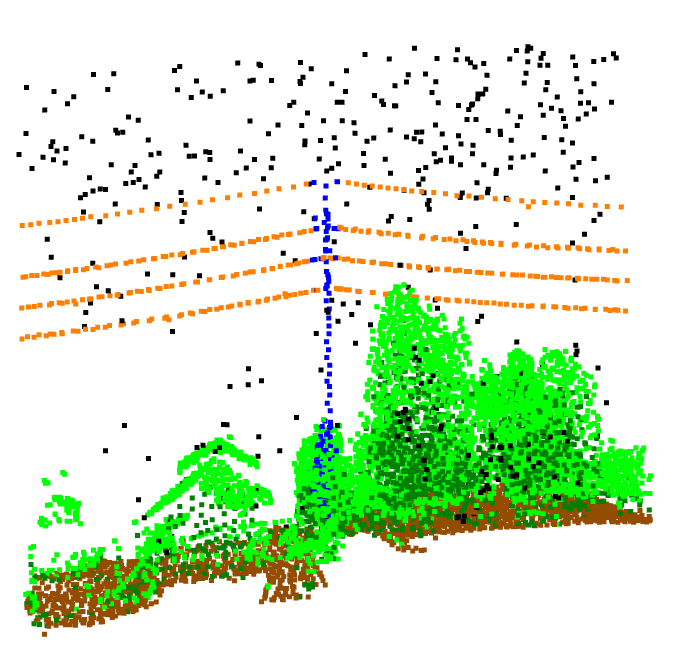} \\
        \includegraphics[width=0.31\columnwidth]{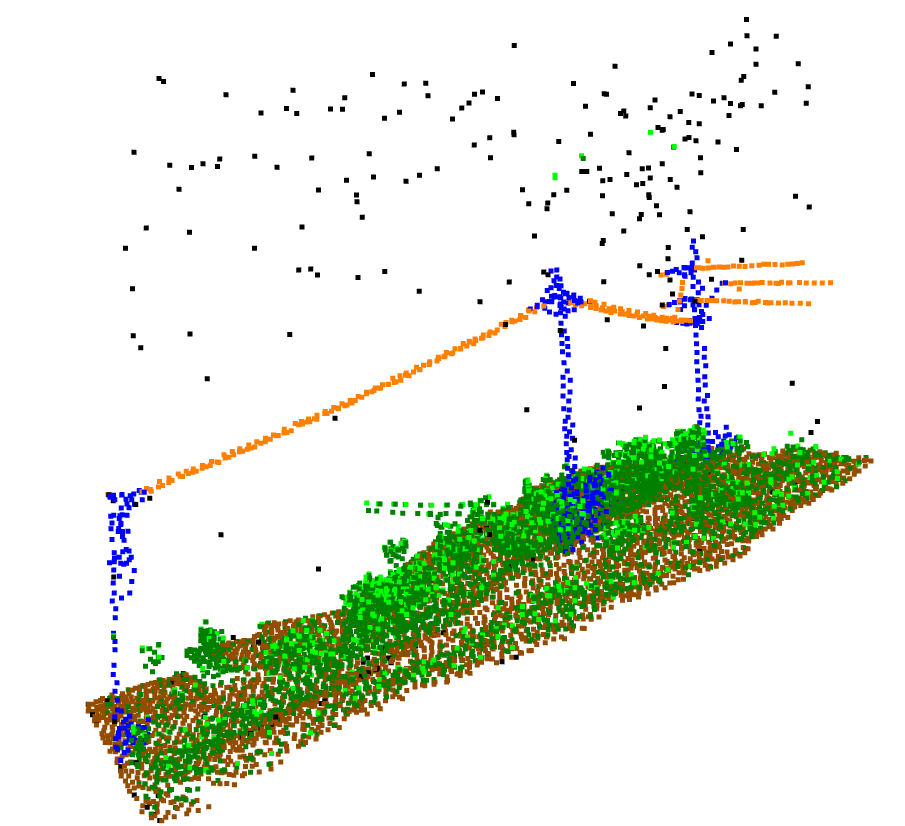} &
        \includegraphics[width=0.31\columnwidth]{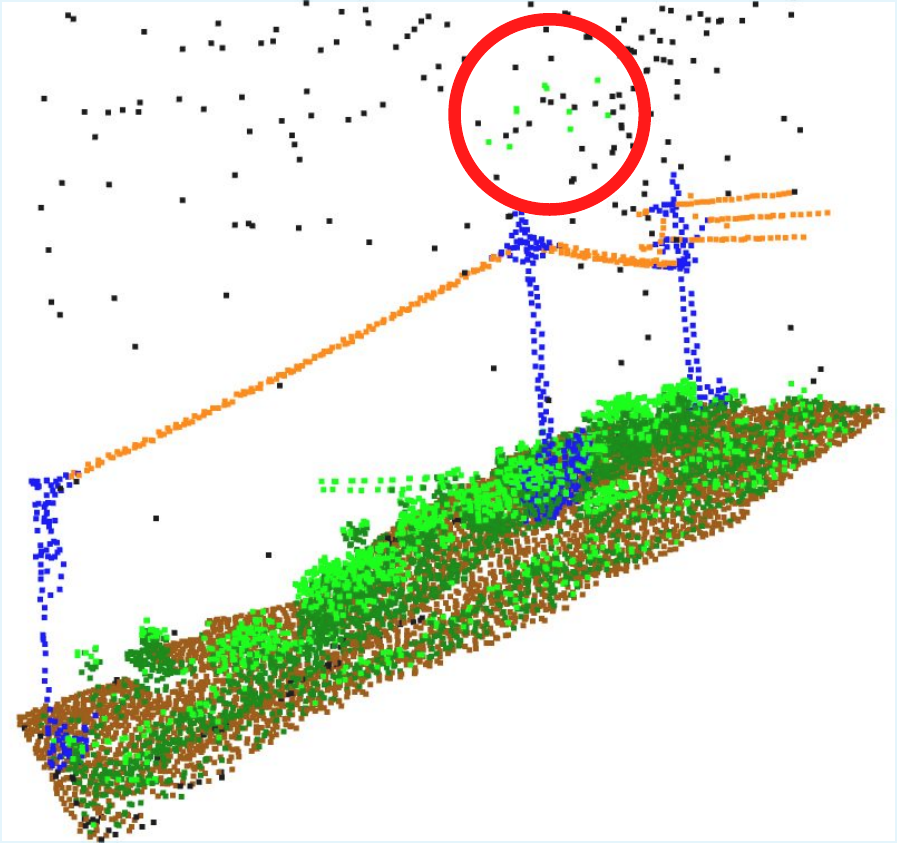} &
        \includegraphics[width=0.31\columnwidth]{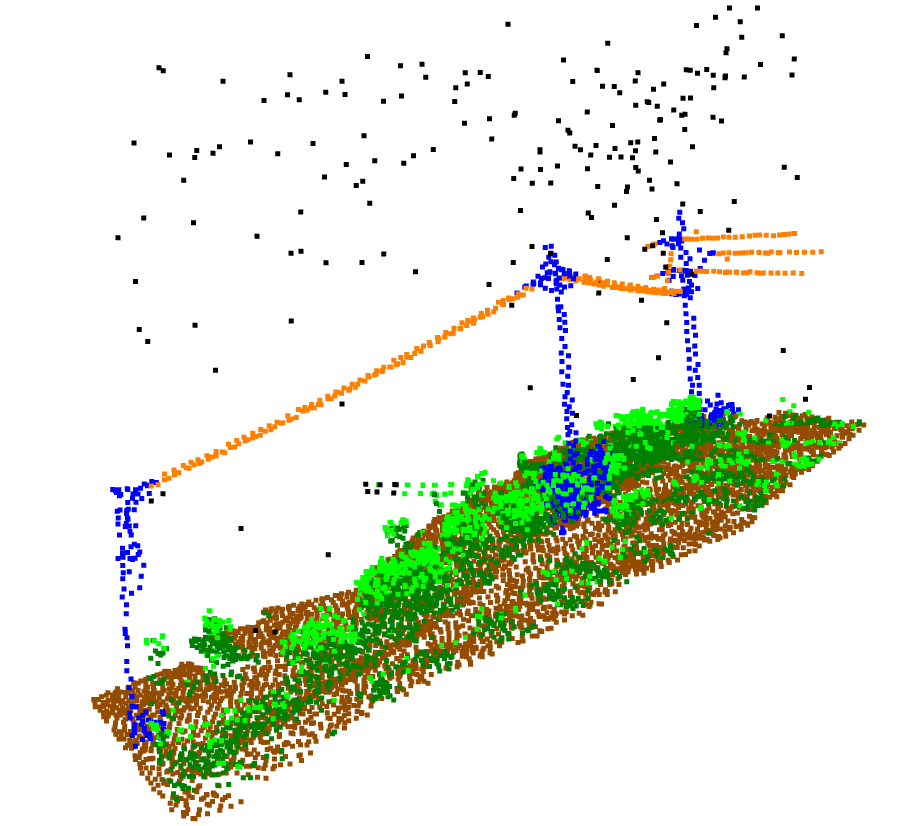} \\
        \multicolumn{3}{c}{
            \includegraphics[width=0.95\columnwidth]{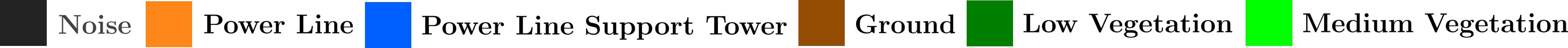}
        } \\
        \multicolumn{1}{c}{\small Input Point Cloud} &
        \multicolumn{1}{c}{\small PTV3 (Baseline)} &
        \multicolumn{1}{c}{\small PTV3 + GIBLy}
    \end{tabular}
    \caption{
    \textbf{Qualitative results on the TS40K dataset.} 
    Each row presents a different scene. From left to right: input point cloud, prediction from the baseline PointTransformerV3 (PTV3), and prediction from PTV3 augmented with GIBLy.  
    In the three rows, the baseline fails to detect support towers, introduces spurious noise in vegetation regions, or misclassifies artifacts as vegetation. In contrast, the GIBLy-augmented model avoids these errors, producing more geometrically consistent predictions. Full view in Supplementary material.
    }
    \label{fig:qual_ts40k}
\end{figure}

\begin{table*}[t]
\centering
\caption{Semantic segmentation results (mIoU \%) and improvements ($\Delta$) across four benchmarks.\textsuperscript{1} Gains are shown in \textcolor{ForestGreen}{green} and deficits in \textcolor{BrickRed}{red}.}
\label{tab:sota-results}
\begin{tabular}{llccc|ccc|ccc|ccc}
\toprule
& & \multicolumn{3}{c|}{\textbf{nuScenes}} & \multicolumn{3}{c|}{\textbf{SemanticKITTI}} & \multicolumn{3}{c|}{\textbf{ScanNet v2}} & \multicolumn{3}{c}{\textbf{S3DIS (Area 5)}} \\
\textbf{Method} & & mIoU & $\Delta$ & & mIoU & $\Delta$ & & mIoU & $\Delta$ & & mIoU & $\Delta$ & \\
\midrule
KPConv & base & 50.74 & -- & & 45.24 & -- & & 36.95 & -- & & 30.31 & -- & \\
& + GIBLy & 37.33 & \textcolor{BrickRed}{--13.41} & & 47.55 & \textcolor{ForestGreen}{+2.31} & & 29.46 & \textcolor{BrickRed}{--7.49} & & 35.71 & \textcolor{ForestGreen}{+5.40} \\
\midrule
PTV1 & base & 66.29 & -- & & 50.17 & -- & & 33.16 & -- & & 47.14 & -- \\
& + GIBLy & 54.38 & \textcolor{BrickRed}{--11.91} & & 53.70 & \textcolor{ForestGreen}{+3.53} & & 39.08 & \textcolor{ForestGreen}{+5.92} & & 51.14 & \textcolor{ForestGreen}{+4.00} \\
\midrule
PTV2 & base & 65.40 & -- & & 54.15 & -- & & 52.05 & -- & & 56.42 & -- \\
& + GIBLy & 70.26 & \textcolor{ForestGreen}{+4.86} & & 55.88 & \textcolor{ForestGreen}{+1.73} & & 55.27 & \textcolor{ForestGreen}{+3.22} & & 65.10 & \textcolor{ForestGreen}{+8.68} \\
\midrule
PTV3 & base & 74.20 & -- & & 55.09 & -- & & 55.64 & -- & & 60.44 & -- \\
& + GIBLy & 69.91 & \textcolor{BrickRed}{--4.29} & & 58.14 & \textcolor{ForestGreen}{+3.05} & & 57.79 & \textcolor{ForestGreen}{+2.15} & & 67.96 & \textcolor{ForestGreen}{+7.52} \\
\bottomrule
\end{tabular}
\vspace{1mm}
\begin{minipage}{\textwidth}
\footnotesize
\textsuperscript{1} For fair evaluation, all models are trained under the same conditions without the use of additional training data or benchmark-specific augmentation pipelines. Some methods (e.g., PTV3~\cite{wu2023ptv3}) report higher performance using extra data. We instead report performance under a controlled and equal setting to ensure valid comparison between each backbone and its GIBLy-augmented variant. Training configurations are fully described in Supplementary Material.
\end{minipage}
\end{table*}

\vspace{-5pt}
\paragraph{TS40K.} Table~\ref{tab:ts40k} presents the per-class and mean IoU results on the TS40K~\cite{lavado2025learning} dataset. We observe that GIBLy consistently improves performance across all backbones. The largest gains are observed for transformer models, with PTV3~\cite{wu2023ptv3} improving from 63.55\% to 75.03\% mIoU (\textcolor{ForestGreen}{+11.48}). The increase is also notable for lightweight MLPs like PointNet, which gains \textcolor{ForestGreen}{7.95} mIoU. However, KPConv~\cite{thomas2019kpconv} shows small regressions, which we attribute to incompatibility between rigid grid-based kernels and soft geometric biases.

\vspace{-8pt}
\paragraph{nuScenes, SemanticKITTI, ScanNet v2, and S3DIS.} 
Table~\ref{tab:sota-results} shows consistent gains across most backbones and datasets. Improvements are especially strong on S3DIS~\cite{armeni20163d}, with boosts of \textbf{+7.52\%} (PTV3~\cite{wu2023ptv3}) and \textbf{+8.68\%} (PTV2~\cite{wu2022point}), confirming GIBLy's effectiveness in structured indoor scenes. SemanticKITTI~\cite{behley2019semantickitti} also benefits, particularly for PointTransformer models, with up to \textbf{+3.53\%} gain on PTV1~\cite{zhao2021point}.
In ScanNet v2~\cite{dai2017scannet}, GIBLy improves all PointTransformer variants, while PointNet-based models~\cite{qi2017pointnet,qi2017pointnet++} degrade likely due to limited feature capacity. Performance on nuScenes~\cite{caesar2020nuscenes} varies: while PointNet, PointNet++ and PTV2 improve, KPConv~\cite{thomas2019kpconv}, PTV1, and PTV3 underperform.
We do not claim new state-of-the-art performance on these benchmarks but demonstrate consistent improvements across strong backbones. Although our reproduced baselines do not reach the exact performance reported in the original papers, GIBLy consistently improves upon them with exactly the same training setting.

\subsection{Ablation Study}
We conduct ablations on the TS40K validation set to evaluate the impact of key GIBLy design choices, including placement strategy, neighborhood count, normalization, composite biases, number of GIBs, and bias type, with additional studies provided in the Supplementary Material. All experiments use PTV3~\cite{wu2023ptv3} as the backbone.

\vspace{-8pt}
\paragraph{GIBLy Placement.}
As shown in Table~\ref{tab:abl-placement}, placing a single GIBLy layer at the input stage achieves the best performance, improving by \textcolor{ForestGreen}{+10.14} mIoU over the baseline. Adding GIBLy at intermediate stages brings limited additional benefit. Interestingly, placing GIBLy layers at every stage leads to a noticeable drop of \textcolor{BrickRed}{--9.39} mIoU. We attribute this degradation to the loss of geometric detail with the use of downsampling at deeper levels.
%
\begin{table}[t]
\centering
\caption{Ablation on GIBLy placement strategy.}
\label{tab:abl-placement}
\begin{tabular}{lcc}
\toprule
\textbf{Configuration} & \textbf{mIoU (\%)} & \textbf{$\Delta$ (\%)} \\
\midrule
Baseline (PTV3) & 65.80 & \\
GIBLy (input only (default)) & 75.94 & \textcolor{ForestGreen}{+10.14} \\
GIBLy (input + intermediate) & 66.18 & \textcolor{ForestGreen}{+0.38} \\
GIBLy (every stage) & 56.41 & \textcolor{BrickRed}{--9.39} \\
\bottomrule
\end{tabular}
\end{table}
\vspace{-8pt}
\paragraph{Neighborhood Count.}
In Table~\ref{tab:abl-ncount}, we evaluate how the number of neighborhood levels affects segmentation accuracy and computational overhead. A single neighborhood radius yields lower performance, while increasing the number of radii improves accuracy up to a point. Using 2 or 3 neighborhoods achieves the best trade-off between accuracy and latency.
\begin{table}[t]
\centering
\caption{Ablation on the number of GIB neighborhood levels.}
\label{tab:abl-ncount}
\begin{tabular}{lccc}
\toprule
\textbf{\# Neighborhoods} & \textbf{mIoU (\%)} & \textbf{Latency (ms)} \\
\midrule
1 & 69.12 & 23.1 \\
2 & 72.41 & 29.2 \\
3 (default) & 75.94 & 37.0 \\
5 & 68.56 & 50.4 \\
\bottomrule
\end{tabular}
\end{table}

\vspace{-8pt}
\paragraph{Number of GIBs.}
In Table~\ref{tab:abl-numgibs}, we vary the number of GIB instances per layer to study their impact on accuracy and runtime. Using two GIBs per shape achieves the best result. Increasing the number beyond this leads to negative returns.

\begin{table}[t]
\centering
\caption{Ablation on the number of GIB instances per layer.}
\label{tab:abl-numgibs}
\begin{tabular}{lccc}
\toprule
\textbf{GIBs per Prior} & \textbf{mIoU (\%)} & \textbf{Latency (ms)} \\
\midrule
2 (default) & 75.94 & 27.0 \\
4 & 73.92 & 32.6 \\
8 & 71.67 & 39.2 \\
16 & 68.84 & 42.4 \\
\bottomrule
\end{tabular}
\end{table}

\vspace{-8pt}
\paragraph{Bias Shape Variants.}
In Table~\ref{tab:abl-shapes}, we isolate the impact of different GIB families. Using the full set of biases outperforms any subset, with radial-only or hollow-only GIBs showing moderate performance. Ellipsoid biases in isolation yield the lowest performance, showing the benefit of combining different priors.

\begin{table}[ht]
\centering
\caption{Ablation on geometric bias types.}
\label{tab:abl-shapes}
\begin{tabular}{lcc}
\toprule
\textbf{Bias Types} & \textbf{mIoU (\%)} & \textbf{$\Delta$ (\%)} \\
\midrule
Radial only (cylinder, disk) & 71.14 & \textcolor{BrickRed}{--4.80} \\
Hollow only & 71.30 & \textcolor{BrickRed}{--4.64} \\
Ellipsoid only & 68.42 & \textcolor{BrickRed}{--7.52} \\
All GIBs (default) & 75.94 & \\
\bottomrule
\end{tabular}
\end{table}

\vspace{-8pt}
\paragraph{Composite Biases.}
In Table~\ref{tab:abl-composites}, we evaluate the role of composite biases. Using none severely limits performance, while increasing their number progressively improves results up to 16 composites. Beyond this point, performance saturates and slightly decreases, suggesting that a moderate number of composites provides the best balance between expressiveness and overfitting.

\begin{table}[ht]
\centering
\caption{Ablation on composite biases.}
\label{tab:abl-composites}
\begin{tabular}{lc}
\toprule
\textbf{Number of Composites} & \textbf{mIoU (\%)} \\
\midrule
0   & 59.60 \\
2   & 66.06 \\
4   & 69.33 \\
8   & 72.14 \\
16 (default) & \textbf{75.94} \\
64  & 73.75 \\
128 & 72.57 \\
\bottomrule
\end{tabular}
\end{table}

\vspace{-8pt}
\paragraph{GIB Normalization.}
Table~\ref{tab:abl-norm} shows the impact of normalization. Without normalization, performance drops significantly as alignment scores are dominated by neighborhood density. Standard z-score normalization partially recovers performance but produces inconsistent semantics, since well-aligned neighbors may be assigned negative values. Our proposed normalization yields the best results, confirming its importance for stable integration of GIBs.

\begin{table}[t]
\centering
\caption{Ablation on GIB normalization.}
\label{tab:abl-norm}
\begin{tabular}{lcc}
\toprule
\textbf{Normalization Type} & \textbf{mIoU (\%)} \\
\midrule
No Normalization            & 62.92 \\
Standardization             & 67.84 \\
GIB Normalization (default) & \textbf{75.94} \\
\bottomrule
\end{tabular}
\end{table}

%% file: sec/5_conclusions.tex
\section{Conclusions}
\label{sec:conclusions}

In this paper we present GIBLy, a geometric inductive bias layer that integrates lightweight, learnable geometric priors into existing 3D learning models. GIBLy is model-agnostic and can be integrated without altering the underlying architecture. By explicitly encoding shape-aware information, GIBLy increases the performance of the majority of baseline models with only 58K extra parameters and a single layer at the input stage.
%
%
As for limitations, GIBLy relies on predefined geometric templates that, while expressive, may not capture irregular structures. An exciting direction for future work is to extend GIBLy with data-driven biases beyond simple primitives with more degrees of freedom.


%% file: sec/6_supp.tex
\section{Geometric Inductive Biases}
\label{sec:gib-defs}

We define all geometric inductive biases (GIBs) used in GIBLy and provide detailed descriptions of their functional forms and intended geometric coverage. Each GIB is implemented as a radial basis function (RBF) with learnable parameters and rotation, returning an alignment score for a given query point $q$ with its local neighborhood.

All operators share a common structure: a local neighborhood $\mathcal{N}_q$ is centered at query point $q$, and all relative neighbor vectors $x - q$ are rotated using a learnable composite rotation matrix $R_{\boldsymbol{\phi}}$ defined by angles $\boldsymbol{\phi} = [\phi_x, \phi_y, \phi_z]$. The rotated projection is:
\begin{equation}
    z_{\boldsymbol{\phi}}(x) = R_{\boldsymbol{\phi}}^\top (x - q)
\end{equation}

The output of each GIB is then computed as $\psi(x)$, which defines the alignment between point $x$ and the learned geometric prior.

\subsection{Cylinder GIB}
The Cylinder GIB models long, tubular structures that are common in both indoor (table legs, poles) and outdoor (tree trunks, signposts) scenes. It uses a Gaussian function centered on the cylindrical axis:
\begin{equation}
\psi_{cy}(x) = \exp\left(-\frac{\|z_{\boldsymbol{\phi}}(x)\|^2}{2r^2}\right)
\end{equation}
Here, $z_{\boldsymbol{\phi}}(x)$ represents the projection of point $x$ into a canonical coordinate system defined by the learned rotation matrix $R_{\boldsymbol{\phi}}$, which aligns the bias with cylindrical structures at learned orientations. The quantity $\|z_{\boldsymbol{\phi}}(x)\|$ corresponds to the radial distance from the cylinder’s central axis.
The parameter $r$ represents the radius of the cylinder and effectively defines the spread of the Gaussian kernel: points closer to the axis receive high alignment scores, while more distant points are attenuated. The shape parameters are $\eta_{cy} = [r, \boldsymbol{\phi}]$.

\paragraph{Hollow Cylinder GIB.}
To model objects represented by surface-only measurements (e.g., pipes, handles, rings), we introduce the Hollow Cylinder GIB:
\begin{equation}
\psi_{hcy}(x) = \exp\left(-\frac{(\|z_{\boldsymbol{\phi}}(x)\| - r)^2}{2t^2}\right)
\end{equation}
This operator peaks when the distance to the central axis equals $r$, effectively suppressing the solid core and outer space. The shell thickness $t$ controls tolerance. Shape parameters: $\eta_{hcy} = [r, t, \boldsymbol{\phi}]$.

\subsection{Cone GIB}
The Cone GIB captures tapering structures like treetops, roofs, or conic sculptures. It adapts both radius and slope to model variation in size and steepness:
\begin{equation}
\psi_{cn}(x) = \exp\left(
-\frac{\|z_{\boldsymbol{\phi}}(x)\|^2}
{2 \left(r h(x) \tan(\beta \pi)\right)^2}
\right)
\end{equation}
Here, $z_{\boldsymbol{\phi}}(x)$ denotes the centered and rotated coordinates of point $x$, and $h(x)$ refers to its vertical height relative to the cone apex. The denominator defines the scale of the Gaussian kernel at height $h(x)$, and is given by:
\[
2 \left( r \cdot h(x) \cdot \tan(\beta \pi) \right)^2
\]
This term dynamically adjusts the radial spread of the bias depending on the height of the point and the angle of the cone.
The parameter $r$ sets the radius of the cone’s base (at $h(x) = 1$), acting as a reference scale. The parameter $\beta \in [0, 0.5)$ controls the steepness of the cone. We constrain $\beta$ to this interval to ensure that $\tan(\beta \pi)$ remains positive and finite, avoiding vertical slopes or undefined behavior. The value of $\tan(\beta \pi)$ grows with $\beta$, meaning that small $\beta$ produces narrow, steep cones, while larger $\beta$ yields wide, flat cones.
By combining $r$, $h(x)$, and $\beta$, the effective "expected distance" from the central axis increases linearly with both height and slope. This results in a variable standard deviation for the Gaussian kernel, producing a larger radial spread as the height increases.
The term $r \cdot h(x) \cdot \tan(\beta \pi)$ thus defines the expected radial distance from the cone’s axis at each height, allowing the GIB to grow smoothly from tip to base.

\paragraph{Hollow Cone GIB.}
The hollow variant isolates shell-like conical structures, useful for sparse surfaces like treetop canopies:
\begin{equation}
\psi_{hcn}(x) = \exp\left(
-\frac{
\left(\|z_{\boldsymbol{\phi}}(x)\| - r h(x) \tan(\beta\pi)\right)^2}
{2t^2}
\right)
\end{equation}
This kernel peaks along the cone surface, with $t$ controlling thickness. Shape parameters: $\eta_{hcn} = [r, \beta, t, \boldsymbol{\phi}]$.

\subsection{Disk GIB}
The Disk GIB captures flat, circular structures such as manholes, ceiling fixtures, or control panels:
\begin{equation}
\psi_{dk}(x) = \exp\left(-\frac{\|z_{\boldsymbol{\phi}}(x)\|^2}{2r^2} \cdot |w - h(x)| \right)
\end{equation}
The scalar $h(x)$ denotes the height of $x$ in the rotated frame, and $w$ is a learnable vertical threshold that enforces a specific width for the disk.
The Gaussian term enforces circular symmetry, while the vertical component $|w - h(x)|$ attenuates points that lie far above or below the disk plane. The combined result is a soft radial surface response. Parameters: $\eta_{dk} = [r, w, \boldsymbol{\phi}]$.

\paragraph{Hollow Disk GIB.}
This variant focuses on circular rims and holes by rewarding points at fixed radial distance from the center:
\begin{equation}
\psi_{hdk}(x) = \exp\left(-\frac{(\|z_{\boldsymbol{\phi}}(x)\| - r)^2}{2t^2} \cdot |w - h(x)| \right)
\end{equation}
Useful for rims, washers, and rings. Parameters: $\eta_{hdk} = [r, t, w, \boldsymbol{\phi}]$.

\subsection{Ellipsoid GIB}
The Ellipsoid GIB generalizes a sphere to ellipsoids of arbitrary shape and orientation, modeled via a precision matrix:
\begin{equation}
\psi_{ellip}(x) = \exp\left(-\frac{x^\top \Lambda(\boldsymbol{\phi}) x}{2}\right)
\end{equation}
Here, $\Lambda(\boldsymbol{\phi})$ is a positive semi-definite matrix encoding orientation and anisotropic scaling. This GIB captures general volumetric shapes such as heads, bodies, rocks. Parameters: $\eta_{ellip} = [\Lambda, \boldsymbol{\phi}]$.

\paragraph{Hollow Ellipsoid GIB.}
To isolate ellipsoidal surfaces and avoid filling the interior, we define:
\begin{equation}
\psi_{hellip}(x) = \exp\left(-\frac{(\sqrt{x^\top \Lambda(\boldsymbol{\phi}) x} - r)^2}{2t^2}\right)
\end{equation}
The Mahalanobis distance defines an implicit ellipsoidal shell. $r$ sets the shell radius, $t$ its thickness. Parameters: $\eta_{hellip} = [\Lambda, r, t, \boldsymbol{\phi}]$.

\subsection{Summary}
GIBs offer interpretable geometric priors with smooth alignment functions. Hollow variants improve robustness to sparsity and incomplete data. All biases are fused with learned features inside the GIBLy module and contribute to precise geometric reasoning in segmentation tasks.

\section{Computation Time Analysis}
\label{sec:comp-time}
A potential concern with GIBLy is that evaluating geometric inductive biases may appear computationally expensive, as it involves applying multiple geometric kernels at every query point. In practice, however, the computation is highly efficient. Each GIB reduces to the evaluation of a simple parametric radial basis function (RBF) over the distance vectors of a given neighborhood. Once neighborhoods are computed, applying GIBs requires only matrix multiplications and point-wise kernel evaluations, both of which are highly parallelizable on GPUs.
To quantify the overhead, we measured the relative computation cost of each operation in GIBLy. The results are summarized in Table~\ref{tab:comp-time}. The majority of the time (over 60\%) is spent on neighborhood construction, which is a common bottleneck across most point-based 3D segmentation methods and not specific to GIBLy. The actual evaluation of GIBs, including rotations, normalization, and basis function computations, accounts for less than 20\% of the total runtime.
Importantly, GIBLy is applied only once at the input stage, where no downsampling has taken place yet. The produced features are then propagated through the backbone, avoiding repeated overhead in deeper layers.

\begin{table}[h]
\centering
\caption{Relative computation time per operation in GIBLy.}
\label{tab:comp-time}
\begin{tabular}{l c}
\toprule
\textbf{Operation in GIBLy}  & \textbf{Compute time ratio} \\
\midrule
Neighborhood Computation     & 64.45\% \\
$R_\phi$ Computation         & 12.11\% \\
GIB Normalization            & 10.19\% \\
GIB Computation              & 7.38\% \\
Composite Bias Computation   & 2.03\% \\
Other operations             & 4.84\% \\
\bottomrule
\end{tabular}
\end{table}

\section{Analysis of PointNet and PointNet++}
PointNet~\cite{qi2017pointnet} and PointNet++~\cite{qi2017pointnet++} are seminal architectures for 3D scene understanding based on shared MLPs. 
In our experiments, these models benefit noticeably from the proposed GIBLy augmentation. On TS40K, GIBLy improves the mIoU of PointNet by $+7.95$ and PointNet++ by $+9.24$. The improvements are especially pronounced in classes such as \textit{Ground} and \textit{Mid Vegetation}. However, we also observe instability in some categories (e.g., \textit{Noise} and \textit{Low Vegetation}), highlighting the sensitivity of earlier architectures to distributional shifts. Across large-scale benchmarks (nuScenes, SemanticKITTI, ScanNet v2, S3DIS), GIBLy consistently provides moderate gains, though certain drops (notably on ScanNet v2) indicate that augmentation alone cannot fully compensate for the representational limitations of these early models.
Overall, these results demonstrate that even simple and relatively shallow backbones benefit from GIBLy, confirming its generality.

\begin{table*}[t]
\centering
\caption{Semantic segmentation results (mIoU \%) across four benchmarks for PointNet/PointNet++.}
\label{tab:sota-pn}
\begin{tabular}{llccc|ccc|ccc|ccc}
\toprule
& & \multicolumn{3}{c|}{\textbf{nuScenes}} & \multicolumn{3}{c|}{\textbf{SemanticKITTI}} & \multicolumn{3}{c|}{\textbf{ScanNet v2}} & \multicolumn{3}{c}{\textbf{S3DIS (Area 5)}} \\
\textbf{Method} & & mIoU & $\Delta$ & & mIoU & $\Delta$ & & mIoU & $\Delta$ & & mIoU & $\Delta$ & \\
\midrule
PointNet & base & 14.56 & -- & & 16.58 & -- & & 17.20 & -- & & 19.46 & -- & \\
& + GIBLy & 18.52 & \textcolor{ForestGreen}{+3.96} & & 19.83 & \textcolor{ForestGreen}{+3.25} & & 13.64 & \textcolor{BrickRed}{--3.56} & & 21.10 & \textcolor{ForestGreen}{+1.64} \\
\midrule
PointNet++ & base & 18.52 & -- & & 23.33 & -- & & 21.55 & -- & & 26.00 & -- \\
& + GIBLy & 21.30 & \textcolor{ForestGreen}{+2.78} & & 24.59 & \textcolor{ForestGreen}{+1.26} & & 14.78 & \textcolor{BrickRed}{--6.77} & & 26.72 & \textcolor{ForestGreen}{+0.72} \\
\bottomrule
\end{tabular}
\end{table*}

\begin{table*}[t]
\centering
\caption{Semantic segmentation results on TS40K for PointNet/PointNet++.}
\label{tab:ts40k-pn}
\begin{tabular}{llccccccc}
\toprule
\textbf{Method} & & \textbf{mIoU (\%)} & Noise & Ground & Low Veg. & Mid Veg. & Tower & Power Line \\
\midrule
PointNet & base & 30.01 & 49.36 & 54.52 & 46.00 & 14.23 & 0.00 & 35.28 \\
         & + GIBLy & 37.96 & 27.13 & 78.06 & 49.42 & 29.60 & 5.04 & 38.51 \\
         & $\Delta$ & \textcolor{ForestGreen}{+7.95} & \textcolor{BrickRed}{--22.23} & \textcolor{ForestGreen}{+23.54} & \textcolor{ForestGreen}{+3.42} & \textcolor{ForestGreen}{+15.37} & \textcolor{ForestGreen}{+5.04} & \textcolor{ForestGreen}{+3.23} \\
\midrule
PointNet++ & base & 45.99 & 59.27 & 59.99 & 54.36 & 14.55 & 22.61 & 78.41 \\
           & + GIBLy & 55.23 & 65.79 & 72.57 & 29.94 & 59.44 & 18.74 & 84.89 \\
           & $\Delta$ & \textcolor{ForestGreen}{+9.24} & \textcolor{ForestGreen}{+6.52} & \textcolor{ForestGreen}{+12.58} & \textcolor{BrickRed}{--24.42} & \textcolor{ForestGreen}{+44.89} & \textcolor{BrickRed}{--3.87} & \textcolor{ForestGreen}{+6.48} \\
\bottomrule
\end{tabular}
\end{table*}

\section{Ablation Studies}

\subsection{Individual Shape Contributions.}
To further analyze the contribution of different geometric primitives, we evaluated each GIB family in isolation. The results are summarized in Table~\ref{tab:abl-individual}.  
Several observations can be made. First, no single primitive is sufficient to capture the diversity of structures present in TS40K. Cylinders and cones, which represent elongated or axially symmetric forms, achieve modest results on their own. Disks perform slightly better, reflecting their ability to approximate planar surfaces. Ellipsoids stand out as the strongest single primitive, consistent with their flexibility in approximating volumetric forms with smooth curvature.  
Second, hollow variants generally outperform their solid counterparts. This is expected given that LiDAR point clouds often provide surface-only information, making shell-like representations more faithful to the data distribution. For example, hollow ellipsoids and hollow disks achieve better performance than their filled versions.  
Finally, the best performance is obtained when combining all GIBs. The combination outperforms the best individual primitive by a large margin (+7.52 mIoU over ellipsoids), highlighting that the different GIBs provide complementary geometric cues. This supports our design choice of using a diverse family of biases and further motivates the introduction of composite biases, which allow the model to learn meaningful mixtures of these primitives.  

\begin{table}[ht]
\centering
\caption{Ablation on individual shape contributions.}
\label{tab:abl-individual}
\begin{tabular}{lc}
\toprule
\textbf{Types of GIBs} & \textbf{mIoU (\%)} \\
\midrule
All GIBs (default) & \textbf{75.94} \\
Cylinder           & 59.83 \\
Ellipsoid          & 68.42 \\
Disk               & 61.43 \\
Cone               & 57.47 \\
Hollow Cylinder    & 60.13 \\
Hollow Ellipsoid   & 63.78 \\
Hollow Disk        & 63.11 \\
Hollow Cone        & 58.62 \\
\bottomrule
\end{tabular}
\end{table}

\section{Training Setup}

\begin{table}[h]
\centering
\caption{Training configuration for indoor and outdoor benchmarks. All models are trained independently (no joint training).}
\label{tab:training-configs}
\resizebox{\columnwidth}{!}{
\begin{tabular}{lcc}
\toprule
\textbf{Setting} & \textbf{Indoor} & \textbf{Outdoor} \\
\midrule
Epochs & 300 & 100 \\
Batch size & 16 & 16 \\
GPU & A100 & A100 \\
Neighborhood size (factor) & 0.4m (2) & 0.4m (2) \\
Point sampling & FPS (100K points) & FPS (100K points) \\
Validation input & Full-res & Full-res \\
Optimizer & AdamW~\cite{loshchilov2017decoupled} & AdamW~\cite{loshchilov2017decoupled} \\
Learning rate & 0.0001 & 0.0001 \\
Loss (weights) &
\begin{tabular}[c]{@{}l@{}}
Focal~\cite{lin2017focal} (0.2) \\
Tversky~\cite{salehi2017tversky} (0.2) \\
Lovász~\cite{berman2018lovasz} (0.8)
\end{tabular}
&
\begin{tabular}[c]{@{}l@{}}
Focal~\cite{lin2017focal} (0.2) \\
Tversky~\cite{salehi2017tversky} (0.2) \\
Lovász~\cite{berman2018lovasz} (0.8)
\end{tabular} \\
Input features & xyz + normals & xyz + normals \\
\bottomrule
\end{tabular}
}
\end{table}

\begin{table}[h]
\centering
\caption{Data augmentation settings used for indoor and outdoor datasets. The parameter $p$ indicates the probability of applying these transformations on a given sample.}
\label{tab:aug-configs}
\begin{tabular}{lcc}
\toprule
\textbf{Augmentation} & \textbf{Indoor} & \textbf{Outdoor} \\
\midrule
Random dropout (ratio 0.2, p=0.2) & \checkmark & -- \\
Rotation z ($\theta \in [-1, 1]$, p=0.5) & \checkmark & \checkmark \\
Random scale (0.9--1.0) & \checkmark & \checkmark \\
Random flip (p=0.5) & \checkmark & \checkmark \\
Jitter ($\sigma=0.005$, clip=0.02) & \checkmark & \checkmark \\
Color jitter (std=0.05, p=0.95) & \checkmark & -- \\
Sphere crop (max=90K pts) & \checkmark & -- \\
Color normalization & \checkmark & -- \\
Coordinate normalization & \checkmark & \checkmark \\
\bottomrule
\end{tabular}
\end{table}

\paragraph{Training configuration.}  
All models are trained independently (i.e., without joint training) on a single NVIDIA A100 GPU with a batch size of 16. Outdoor datasets are trained for 100 epochs, while indoor datasets use 300 epochs. 
During training, we sample up to 100{,}000 points per scene using farthest point sampling (FPS) to ensure geometric consistency. For evaluation, full-resolution point clouds are used without subsampling. 
Transformer-based frameworks often benefit from modern training strategies that incorporate advanced data augmentations~\cite{qian2022pointnext,wu2022point}.  
Since our method does not rely on these specific optimizations, we retrain state-of-the-art models under the same conditions to ensure a fair comparison across all benchmarks.  
For optimization, we adopt AdamW~\cite{loshchilov2017decoupled}, a widely used optimizer in recent 3D learning methods.  
Following prior works~\cite{zhao2021point,wu2022point,wu2023ptv3,peng2024oa}, we estimate normal vectors for points and include coordinates as additional input features.
All details are summarized in~\cref{tab:training-configs}.

\paragraph{Data Augmentation.}
To encourage robustness and generalization, we adopt a consistent set of data augmentation strategies across all models, as summarized in Table~\ref{tab:aug-configs}. These include random dropout, rotation, jitter, normalization, and color jittering.

\begin{figure*}[ht]
    \centering
    \setlength{\tabcolsep}{1pt}
    \renewcommand{\arraystretch}{1.0}
    \begin{tabular}{ccc}
        \includegraphics[width=0.31\textwidth]{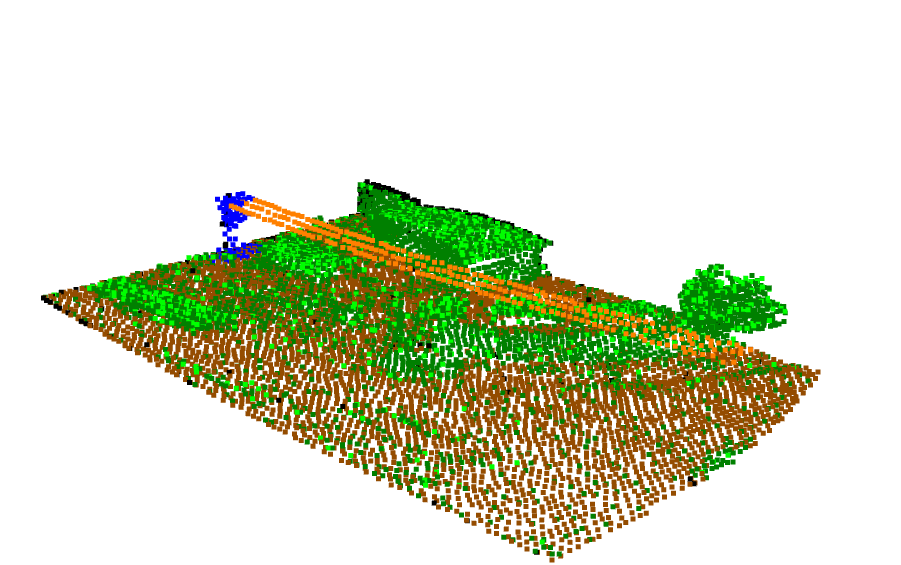} &
        \includegraphics[width=0.31\textwidth]{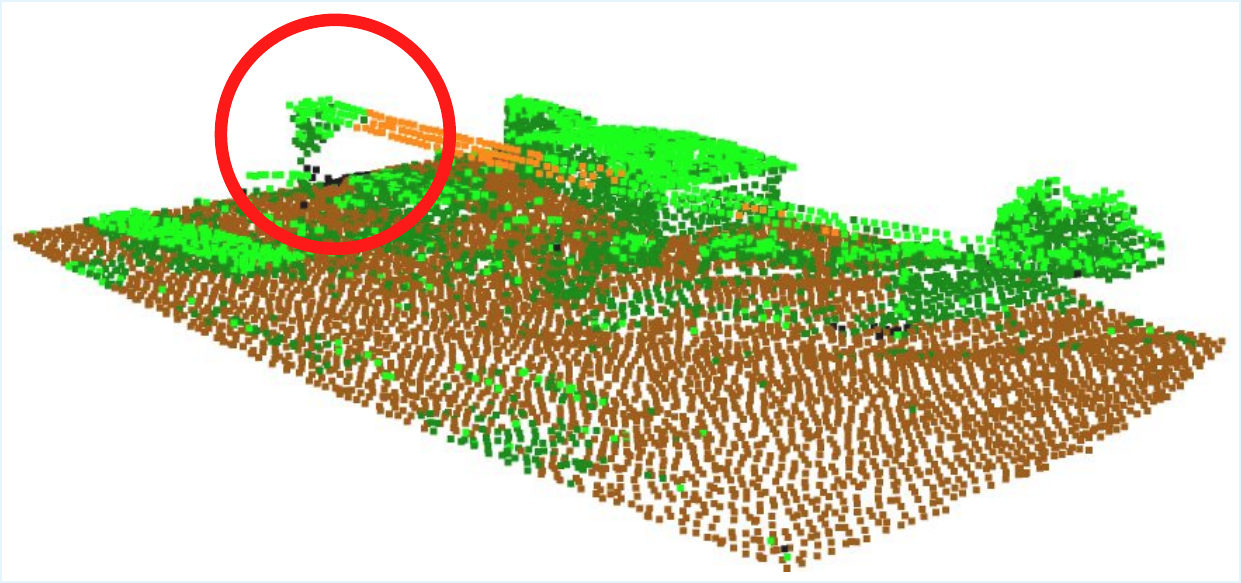} &
        \includegraphics[width=0.31\textwidth]{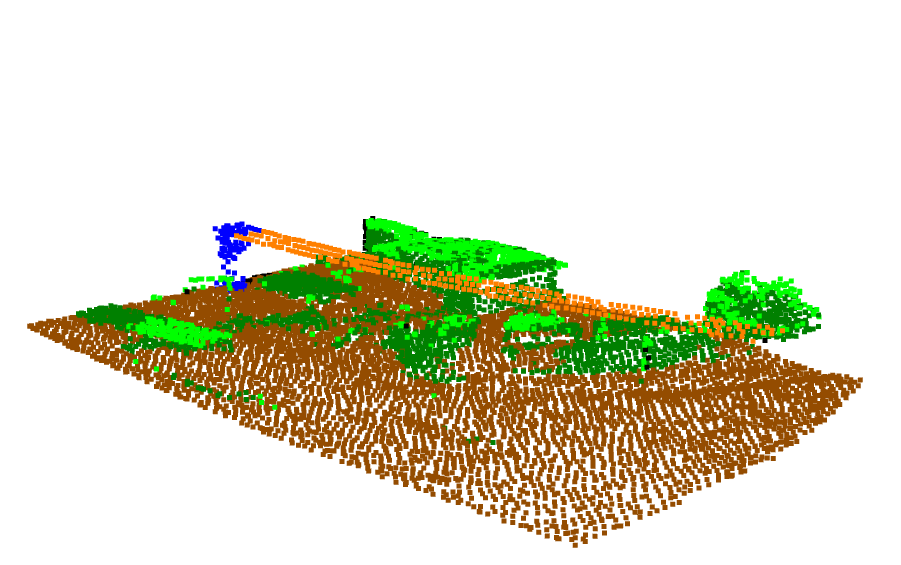} \\
        \includegraphics[width=0.31\textwidth]{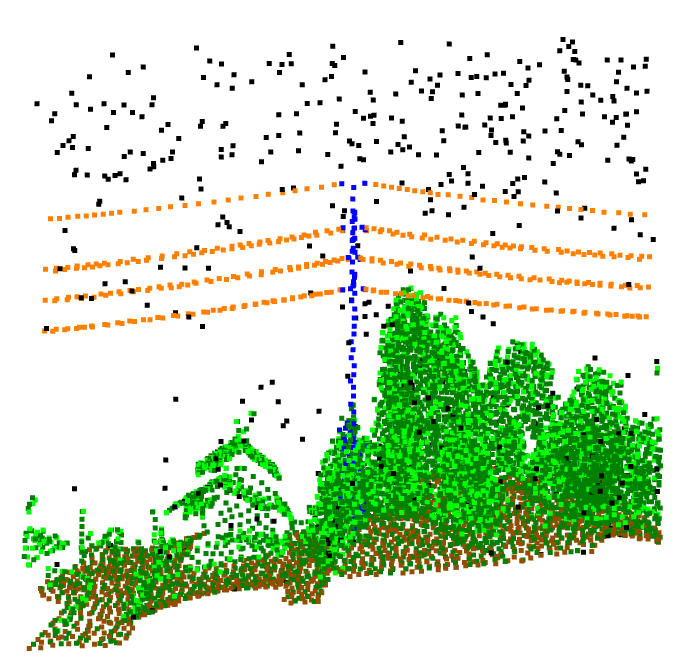} &
        \includegraphics[width=0.31\textwidth]{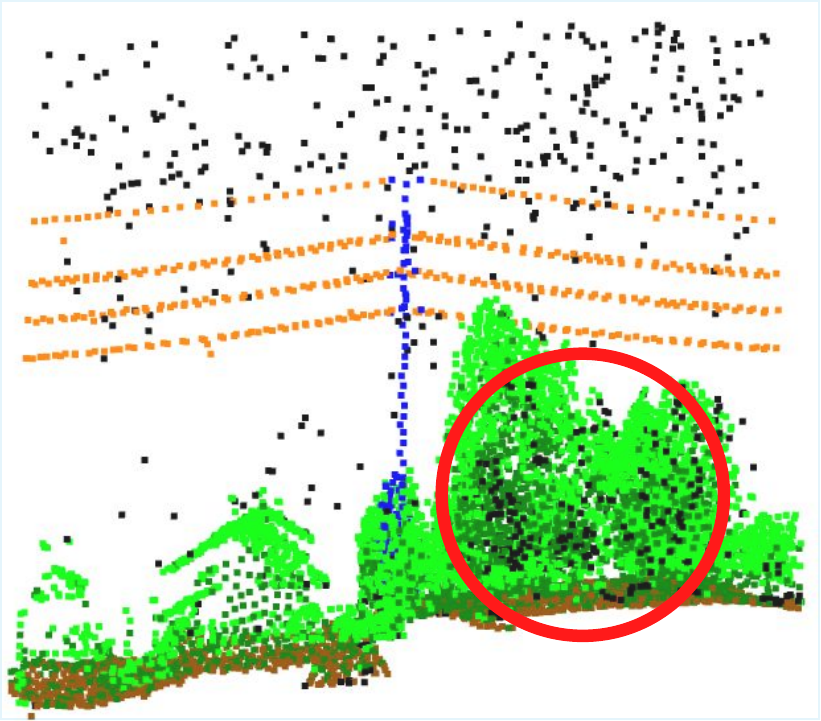} &
        \includegraphics[width=0.31\textwidth]{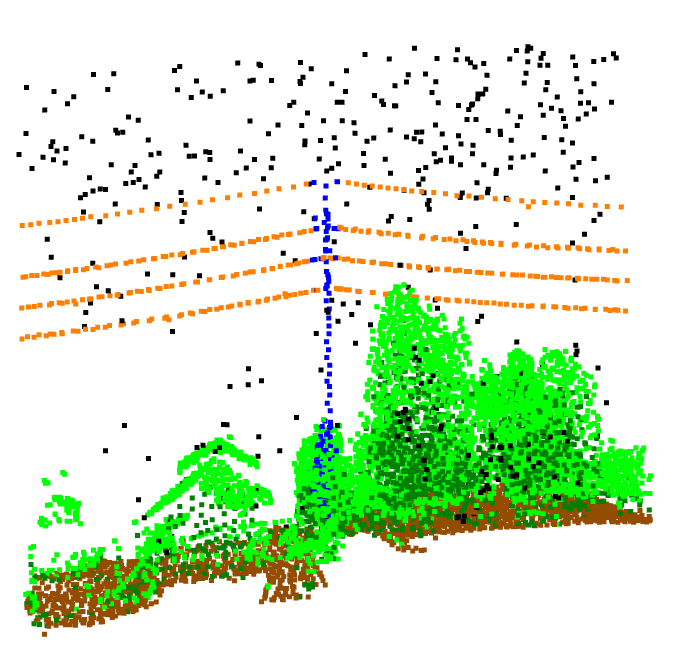} \\
        \includegraphics[width=0.31\textwidth]{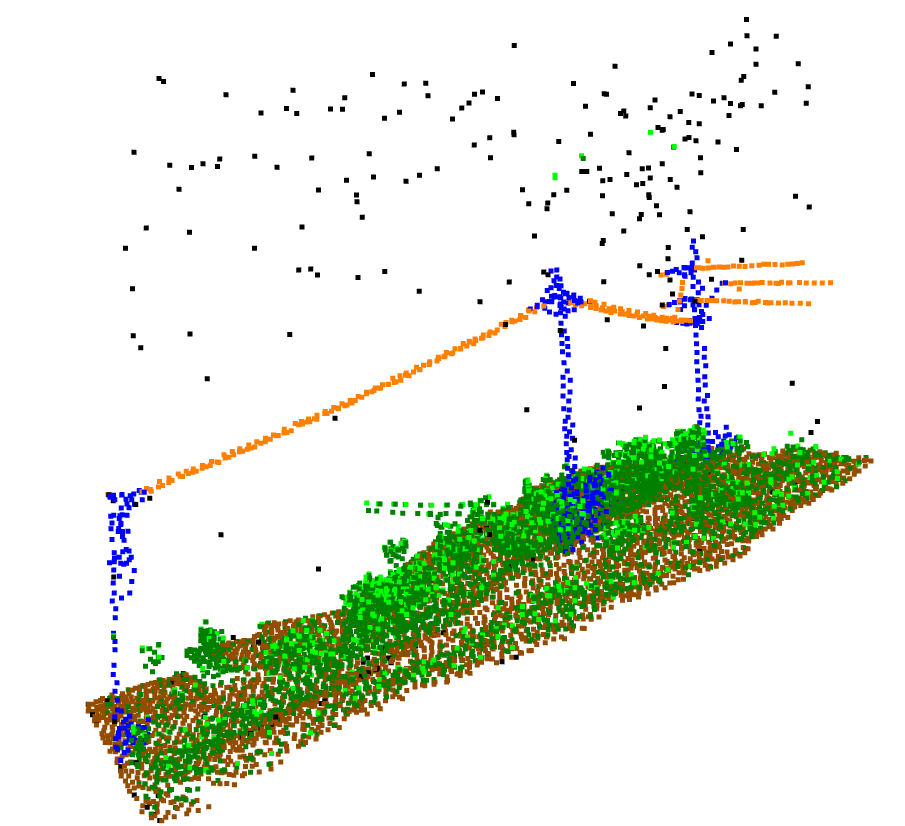} &
        \includegraphics[width=0.31\textwidth]{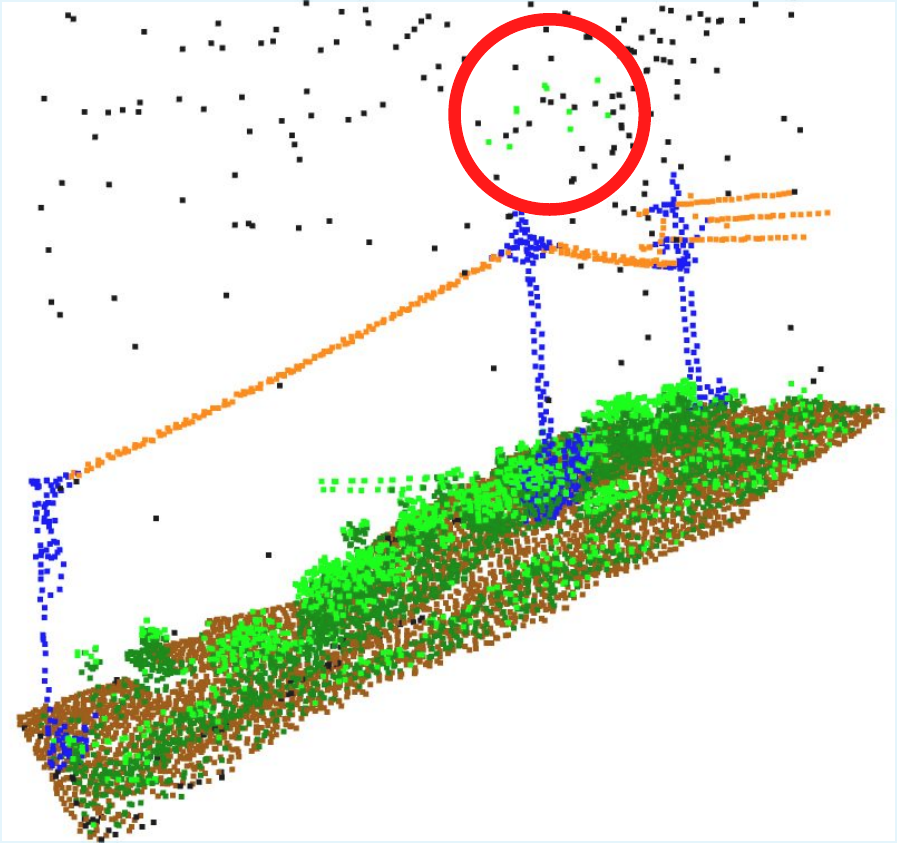} &
        \includegraphics[width=0.31\textwidth]{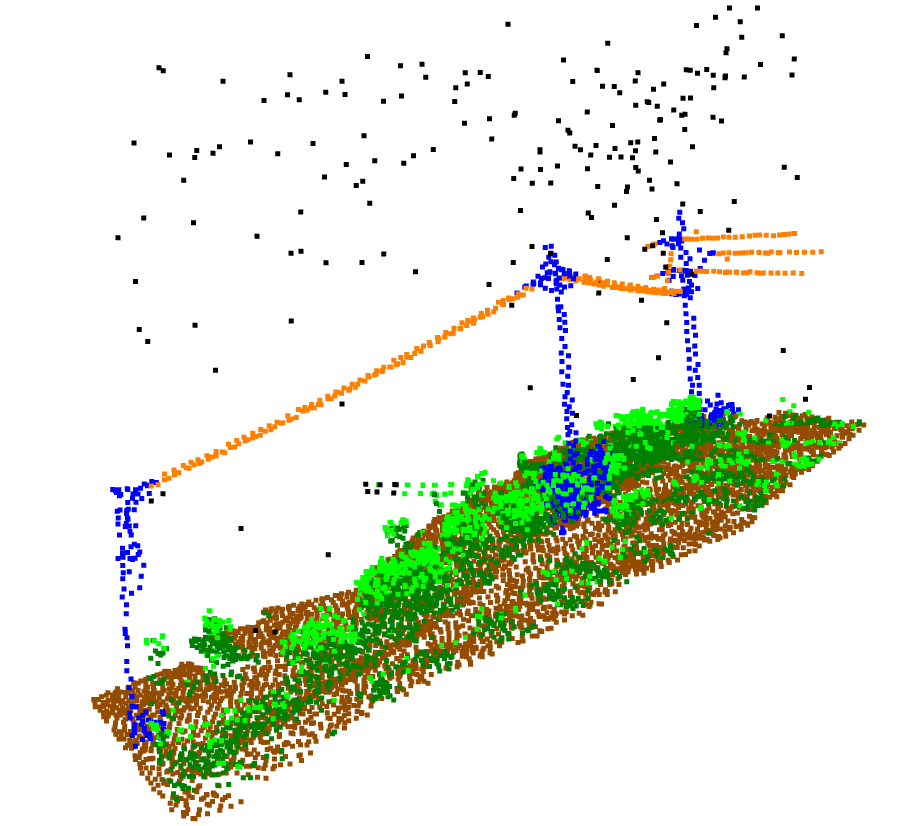} \\
        \multicolumn{3}{c}{
            \includegraphics[width=0.95\textwidth]{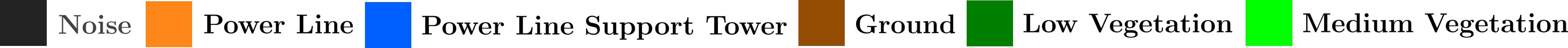}
        } \\
        \multicolumn{1}{c}{\small Input Point Cloud} &
        \multicolumn{1}{c}{\small PTV3 (Baseline)} &
        \multicolumn{1}{c}{\small PTV3 + GIBLy}
    \end{tabular}
    \caption{
    \textbf{Qualitative results on the TS40K dataset.} 
    Each row presents a different scene. From left to right: input point cloud, prediction from the baseline PointTransformerV3 (PTV3), and prediction from PTV3 augmented with GIBLy.  
    In the three rows, the baseline fails to detect support towers, introduces spurious noise in vegetation regions, or misclassifies artifacts as vegetation. In contrast, the GIBLy-augmented model avoids these errors, producing more geometrically consistent predictions.  
    }
    \label{fig:supp_qual}
\end{figure*}

\paragraph{Qualitative Results.} We provide a full view of qualitative results in Figure~\ref{fig:supp_qual} to illustrate the effect of GIBLy across diverse 3D scenes. Each row compares the baseline prediction against its GIBLy-augmented version for a different sample. As observed, GIBLy enhances structural consistency and reduces misclassifications in geometrically complex regions.

